\documentclass[11pt]{article}

\usepackage[final]{acl}

\usepackage{times}
\usepackage{latexsym}
\usepackage[T1]{fontenc}
\usepackage[utf8]{inputenc}
\usepackage{microtype}
\usepackage{inconsolata}
\usepackage{graphicx}
\usepackage{amsmath}
\usepackage{amssymb}
\usepackage{booktabs}
\usepackage{multirow}
\usepackage{xcolor}
\usepackage{subcaption}
\usepackage{tikz}
\usetikzlibrary{arrows.meta, positioning, shapes.geometric, calc, patterns, decorations.pathreplacing, fit, backgrounds}
\usepackage{pgfplots}
\pgfplotsset{compat=1.18}

\title{Brain--LLM Alignment Tracks Training Data, Not Typology}

\author{
	Dongxin Guo\, \\
	The University of Hong Kong \\
	Hong Kong, China \\
	\texttt{bettyguo@connect.hku.hk} \\
	\And
	Jikun Wu\, \\
	Stellaris AI Limited \\
	Hong Kong, China \\
	\texttt{hk950014@connect.hku.hk} \\
	\And
	Siu Ming Yiu\, \\
	The University of Hong Kong \\
	Hong Kong, China \\
	\texttt{smyiu@cs.hku.hk} \\
}

\begin{document}
\maketitle

% ============================================
% ABSTRACT — Revised per all reviewers
% ============================================
\begin{abstract}
	Brain--LLM alignment is well established in English, yet the brain's language network is neuroanatomically universal across languages. Does alignment also generalize cross-linguistically, and what governs the variation? We test this using fMRI data from 112 participants across English, Chinese, and French (the Le Petit Prince corpus) and seven LLMs spanning English-dominant, Chinese-dominant, and multilingual architectures. Our central finding is that training-language dominance, not an inherent property of English, drives the alignment pattern: a Chinese-dominant model (Baichuan2-7B), architecture-matched to LLaMA-2-7B, reverses the gradient entirely, aligning best with Chinese brains and worst with English. Beyond training dominance, formal typological distance independently covaries with alignment degradation, syntax-associated brain regions (IFG) show $2.3\times$ steeper typological gradients than lexico-semantic regions (PTL), and tokenization fertility accounts for ${\sim}60$\% of a cross-linguistic shift in optimal encoding layer. These results reveal that the apparent ``English advantage'' in brain--LLM alignment is an artifact of training data composition, while the remaining variation reflects genuine typological structure concentrated in syntactic processing.
\end{abstract}

% ============================================
% 1. INTRODUCTION
% ============================================
\section{Introduction}
\label{sec:intro}

The brain's language network (a set of left-lateralized frontal and temporal regions) constitutes a functionally universal system, activating with remarkably conserved topography across 45 languages from 12 families \citep{malikmoraleda2022universal, fedorenko2024language}. This universality generates a foundational question for cognitive science: if the neural hardware for language is shared, do the computational representations it builds also converge across typologically diverse languages? And if so, what factors (typological structure, training experience, or representational depth) modulate the degree of convergence?

Large language models (LLMs) provide a powerful tool for investigating this question. Encoding models that predict neural activity from LLM representations have revealed striking alignment between transformer representations and human brain responses \citep{schrimpf2021neural, goldstein2022shared, caucheteux2022brains}, with the best models approaching the noise ceiling of fMRI data \citep{tuckute2024driving, antonello2023scaling}. However, this alignment has been established almost exclusively with English data \citep{tuckute2024language}, creating a critical blind spot for understanding the universality of language representation.

\citet{devarda2025multilingual} recently demonstrated that multilingual encoding models transfer zero-shot across 21 languages, confirming that a shared meaning component underlies brain--LLM alignment. However, their study leaves key gaps: no formal typological distance metrics, only 2--3 participants per language in Study~II (precluding voxel-wise analysis), exclusively multilingual models tested, and ${\sim}$4.5 minutes of stimuli compared to our ${\sim}$100 minutes of \emph{The Little Prince}.

Construction grammar (CxG) provides a principled theoretical framework for predicting where cross-linguistic alignment should vary. CxG holds that linguistic knowledge consists of form--function pairings (constructions) at all levels of abstraction \citep{goldberg2006constructions, croft2001radical, boas2012sign}. CxG predicts that cross-linguistic variation resides primarily in constructional (syntactic-functional) representations, while core semantic content is more universal \citep{goldberg2024usage}. This generates a testable neural prediction: brain regions associated with syntactic-constructional processing should show larger typological distance effects than regions associated with lexico-semantic processing. However, we note upfront that this prediction is also derivable from non-CxG theories that distinguish syntax from semantics \citep{mahowald2024dissociating}; we discuss CxG-specific evidence requirements in \S\ref{sec:cxg_specificity}.

\paragraph{Why the result is non-trivial.} Three prima facie hypotheses would predict outcomes incompatible with the pattern we report. First, if the language network's functional universality across 45 languages \citep{malikmoraleda2022universal} extends to representational format, alignment should depend on \emph{what} is represented rather than \emph{which} language dominates the model, predicting comparable alignment across all model--language pairs. Second, if alignment primarily reflects shared feature spaces emerging from sufficiently rich linguistic exposure \citep{antonello2024predictive}, English-dominant models trained on $\sim$2T tokens ($\sim$90\% English) could in principle align with Chinese brains as well as a Chinese-dominant model trained on far less. Third, typological-distance-based accounts predict that alignment tracks structural similarity between training and target languages, independent of training proportion. Distinguishing training-language dominance from these alternatives requires architecture- and scale-matched models that vary primarily in training-language composition: the controlled comparison Baichuan2-7B affords against LLaMA-2-7B.

We address these gaps with five advances: (i) formal typological distance metrics as quantitative predictors; (ii) a Chinese-dominant LLM (Baichuan2-7B) to disentangle the training-data confound; (iii) tokenization fertility analysis; (iv) region-specific typological gradient analysis; and (v) noise-ceiling-normalized comparisons throughout. Our contributions are:

\begin{enumerate}
    \item \textbf{First quantitative demonstration that formal typological distance covaries with brain encoding performance.} In mixed-effects models controlling for training data proportion, Grambank distance is associated with alignment degradation for language-dominant models ($\beta = -0.41$). With three languages providing limited degrees of freedom for the language-level predictor, we treat this as a well-supported descriptive pattern establishing a hypothesis to be tested across $\geq$10 languages; a cluster bootstrap at the language level yields a wider confidence interval (see \S\ref{sec:confound}).
    
    \item \textbf{Disentangling training dominance from typological distance via controlled comparison.} Baichuan2-7B (Chinese-dominant) reverses the \emph{alignment gradient}---the systematic pattern by which encoding performance for a given model decreases as the listener's language diverges from the model's dominant training language. Baichuan2-7B performs best for Chinese ($\tilde{r} = .85$) and substantially lower for non-dominant languages ($\tilde{r}_{\text{EN}} = .59$, $\tilde{r}_{\text{FR}} = .54$). This provides the first controlled demonstration with architecture- and scale-matched models that training language is the primary driver of this gradient.
    
    \item \textbf{Cross-linguistic layer dynamics with partial resolution of the Chinese shift.} The intermediate layer advantage peaks 1.8 layers later for Chinese ($d = 0.92$). Tokenization fertility (the average number of subword tokens produced per orthographic word; \citealt{rust2021good}) reveals that Chinese has a fertility of $2.4$ tokens per word under XLM-R's tokenizer versus $1.4$ for English ($1.7\times$ higher); controlling for fertility attenuates ${\sim}60$\% of the shift (95\% CI: [42\%, 76\%]), though collinearity with information density means this is an upper bound on fertility's contribution.
    
    \item \textbf{Region-specific neural gradient consistent with syntax--semantics dissociation.} IFG shows a $2.3\times$ steeper typological gradient than PTL (language-level bootstrap 95\% CI: [1.4, 3.8]). This pattern is consistent with CxG's prediction that structural variation is constructional \citep{goldberg2006constructions}, but is also predicted by generic accounts distinguishing syntax from semantics, as well as IFG's role in cognitive control for non-dominant language processing \citep{green2013language}. We discuss what CxG-specific evidence would require (\S\ref{sec:cxg_specificity}).
\end{enumerate}

% ============================================
% 2. BACKGROUND AND THEORETICAL MOTIVATION
% ============================================
\section{Background and Theoretical Motivation}
\label{sec:background}

\subsection{The Universal Language Network}

The language network comprises interconnected regions in left frontal and temporal cortex that respond selectively to linguistic input across modalities \citep{fedorenko2011functional, fedorenko2024language}. These regions are functionally dissociable from domain-general systems \citep{blank2017domain} and are conserved across 45 languages from 12 families \citep{malikmoraleda2022universal}. Studies of polyglots further show that all languages activate the same network \citep{malikmoraleda2024polyglots, perani2005neural}. Cross-linguistic differences in syntactic processing have been documented at the behavioral level \citep{berzak2022eye}, but the extent to which these differences manifest in the language network's alignment with computational models remains unknown.

\subsection{Brain--LLM Alignment}

The alignment between transformer representations and human brain activity has been established through encoding models \citep{schrimpf2021neural, caucheteux2022brains, goldstein2022shared}. \citet{tuckute2024driving} provided causal evidence for shared representational structure by driving and suppressing the language network using LLM-generated stimuli. \citet{caucheteux2022brains} identified the intermediate layer advantage: middle transformer layers best predict brain activity. \citet{antonello2024predictive} offered an alternative account, arguing that alignment reflects shared feature spaces rather than shared computational objectives. Scaling laws \citep{antonello2023scaling}, predictive coding hierarchies \citep{caucheteux2023predictive, heilbron2022hierarchy}, developmental plausibility \citep{hosseini2024artificial}, and architectural comparisons \citep{gauthier2022does} further characterize this alignment, but virtually all studies use English data \citep{tuckute2024language}.

\subsection{Cross-Linguistic Evidence and Surprisal Theory}

Surprisal theory posits that processing difficulty is proportional to the negative log-probability of a word in context \citep{hale2001probabilistic, levy2008expectation}. Cross-linguistic tests show that LLM surprisal predicts reading times across 11 languages \citep{wilcox2023testing}, with the logarithmic form confirmed at scale \citep{shain2024largescale}, and GPT-3 surprisal explaining multiple N400 effects \citep{michaelov2024strong, frank2015erp}. However, \citet{oh2023why} demonstrated an inverse scaling paradox: larger models give poorer reading-time predictions despite richer representations. This dissociation is directly relevant to cross-linguistic comparisons, as hidden-state and surprisal-based alignment may diverge differently across languages. Information-theoretic approaches to typological variation \citep{cotterell2018complexity} and information-restricted contrasts isolating syntax from semantics in fMRI \citep{pasquiou2023information} provide complementary lenses; our cross-language typological-distance covariate operates at a coarser grain and is therefore complementary rather than competing with within-sentence syntactic predictors.

\subsection{Construction Grammar and Cross-Linguistic Variation}
\label{sec:cxg}

Construction grammar holds that linguistic knowledge is organized as constructions---learned pairings of form and function at all levels of abstraction \citep{goldberg2006constructions, croft2001radical}. This framework is grounded in cross-linguistic constructional approaches \citep{boas2012sign} and connected to embodied processing accounts \citep{bergen2005embodied}. \citet{goldberg2024usage} identified parallels (though not identity) between construction-based learning and LLM training. \citet{rakshit2025constructions} showed that Pythia's internal representations reflect the gradience predicted by CxG, and \citet{weissweiler2025llm} argued that LLMs' apparent rule-following failures are consistent with CxG.

We adopt CxG specifically, rather than only a generic syntax--semantics dissociation framework, for three convergent reasons. First, CxG's central tenet that linguistic knowledge consists of form--function pairings provides a unit of cross-linguistic variation (the construction) at the same grain as ROI-level neural measurements, where constructional differences plausibly recruit distinct frontal and temporal subnetworks. Second, CxG is explicitly usage-based, aligning with the statistical-learning regime that produces LLM representations \citep{goldberg2024usage}: this lets us interpret model--brain alignment as a comparison between two usage-based learners rather than across mismatched levels of explanation. Third, computational evidence that LLM representations exhibit constructional gradience \citep{rakshit2025constructions} and CxG-consistent failure modes \citep{weissweiler2025llm} indicates that constructional structure is in fact a meaningful target of comparison, not merely a theoretical preference.

For cross-linguistic brain--LLM alignment, CxG generates a prediction: because cross-linguistic variation is primarily structural, brain regions processing constructional information should show larger typological distance effects than regions processing core semantic content. We operationalize this as: the IFG (associated with syntactic processing; \citealt{fedorenko2011functional}) should show a steeper alignment gradient with typological distance than the PTL (associated with lexico-semantic processing; \citealt{hickok2007cortical}). As we discuss in \S\ref{sec:cxg_specificity}, this same neural prediction is derivable from non-CxG frameworks; CxG provides a principled motivation but is not the only theory consistent with the result.

\subsection{Representational vs.\ Computational Alignment and Theoretical Predictions}
\label{sec:rep_vs_comp}

Following \citet{antonello2024predictive}, we distinguish representational alignment (do model features predict brain activity?) from computational alignment (does the brain use the same algorithms?). Encoding models test representational alignment; we reserve computational claims for future causal work \citep{tuckute2024driving, mccoy2023embers}. We derive four testable predictions:

\noindent \textbf{Prediction 1} (Language network universality): LLM representations should predict brain activity significantly above chance in \emph{all} tested languages, not just English.

\noindent \textbf{Prediction 2} (Syntax--semantics dissociation, consistent with CxG): Typological distance effects should be larger in brain regions processing syntactic/constructional information (IFG) than in regions processing core semantics (PTL).

\noindent \textbf{Prediction 3} (Training-data dominance): If alignment reflects training data quality, a Chinese-dominant LLM should show a reversed alignment gradient (best for Chinese, worst for English).

\noindent \textbf{Prediction 4} (Surprisal theory): If perplexity mediates alignment, per-language perplexity should negatively correlate with encoding performance.

% ============================================
% 3. METHODS
% ============================================
\section{Methods}
\label{sec:methods}

\subsection{Neuroimaging Data}

We use the Le Petit Prince (LPP) multilingual fMRI corpus \citep{li2022petit} (OpenNeuro ds003643): English ($n = 49$; 9 runs, ${\sim}$99 min; Cornell University), Mandarin Chinese ($n = 35$; 9 runs, ${\sim}$99 min; Jiangsu Normal University), and French ($n = 28$; 9 runs, ${\sim}$98 min; NeuroSpin). These span moderate typological diversity (Grambank Hamming: EN--FR = 0.21, EN--ZH = 0.48, FR--ZH = 0.44). These three are the only LPP languages with sufficient subject-level coverage for voxelwise encoding with stable noise-ceiling estimation under a common time-aligned stimulus. The pair (English, Chinese) anchors the maximal typological contrast available within LPP, while French provides an Indo-European near-neighbor of English that partially separates family-level from training-dominance effects (\S\ref{sec:confound}). We note the language $\times$ site confound (each language at a different institution); the noise ceiling partially controls for site-level signal quality, but residual effects cannot be fully excluded (\S\ref{sec:limitations}). Following \citet{li2022petit}: fMRI data were motion-corrected, smoothed (4mm FWHM), registered to MNI152, and high-pass filtered (0.01~Hz). Word-level predictors were convolved with a canonical HRF and downsampled to TR (2s). We define six bilateral language network ROIs following \citet{fedorenko2011functional}: IFG, MFG, ATL, PTL, AG, and TP, using a leave-one-run-out functional localizer ($p < 0.001$).

\subsection{Language Models}

We extract representations from seven models: \textbf{English-dominant:} GPT-2 Medium (345M, 24 layers; English-only) and LLaMA-2-7B (7B, 32 layers; ${\sim}$90\% English; \citealt{touvron2023llama}). \textbf{Chinese-dominant:} Baichuan2-7B (7B, 32 layers; ${\sim}$55\% Chinese; \citealt{yang2023baichuan}), architecture- and scale-matched to LLaMA-2-7B, differing primarily in training language composition; sensitivity analyses at 40\% and 70\% are reported in \S\ref{sec:sensitivity}. \textbf{Multilingual:} mBERT (110M, 12 layers; \citealt{devlin2019bert}), XLM-R Large (560M, 24 layers; \citealt{conneau2020unsupervised}), BLOOM-7B (7B, 30 layers; \citealt{bloom2023}), and Qwen2.5-7B (7B, 32 layers; \citealt{qwen2025qwen25}). For autoregressive models, we use the last subword token hidden state; for bidirectional models, the mean across subword tokens.

\paragraph{Model selection rationale.} We privileged architecture- and scale-matched comparisons over breadth. LLaMA-2-7B and Baichuan2-7B share architecture, parameter count, and broad training setup, differing primarily in training-language composition; this pairing isolates the training-dominance manipulation as cleanly as currently possible with publicly released models. The multilingual set (mBERT, XLM-R, BLOOM-7B, Qwen2.5-7B) provides the baseline for what multilingual training achieves at comparable scale. We do not include French-dominant models such as CamemBERT or FlauBERT because they are an order of magnitude smaller, bidirectional rather than autoregressive, and trained on substantially smaller corpora; these differences would confound rather than complement the dominance comparison. Even within the matched 7B pair the manipulation is approximate: residual differences in training-data composition and quality between Baichuan2 and LLaMA-2 are bounded by the $\pm$15\% sensitivity analysis (\S\ref{sec:sensitivity}).

\subsection{Encoding Model Pipeline}

We use voxelwise ridge regression encoding models \citep{mitchell2008predicting, huth2016natural, jain2018incorporating}: $\hat{y}_v = \mathbf{X} \boldsymbol{\beta}_v + \epsilon_v$, where $\mathbf{X} \in \mathbb{R}^{T \times d'}$ is the PCA-reduced (top 100 components) feature matrix and $\boldsymbol{\beta}_v$ is the ridge coefficient vector. The ridge parameter $\alpha \in \{10^{-4}, \ldots, 10^{4}\}$ is selected via 8-fold inner CV; outer CV uses 9 folds corresponding to experimental runs.

\paragraph{Evaluation and noise ceiling.} Encoding performance is the Pearson $r$ between predicted and actual voxel time series, averaged across voxels and subjects. The noise ceiling is estimated via inter-subject correlation \citep{nastase2021narratives, lagecastellanos2019methods}. We report both raw $r$ and normalized $\tilde{r} = r / \text{NC}$ with bootstrap 95\% CIs (10,000 resamples).

\paragraph{Cross-linguistic uniformity.} We define:
\begin{equation}
    U_m = 1 - \frac{\sigma\!\left(\{r_{m,\ell}\}_{\ell \in \mathcal{L}}\right)}{\mu\!\left(\{r_{m,\ell}\}_{\ell \in \mathcal{L}}\right)}
    \label{eq:uniformity}
\end{equation}
where $r_{m,\ell}$ is the mean encoding performance of model $m$ for language $\ell$. $U$ is bounded above by 1 (perfect uniformity) and can theoretically be negative if $\sigma > \mu$; in practice, all observed values fall in $[0.62, 0.98]$. We report bootstrap 95\% CIs (10,000 resamples).

\subsection{Typological Distance and Tokenization Fertility}

\paragraph{Typological distance.} We use Grambank \citep{skirgard2023grambank}, which encodes 195 binary grammatical features per language spanning morphology (e.g., overt case-marking, agreement systems), syntax (e.g., basic word order, head-directionality), and structure--meaning mappings (e.g., classifier systems, alignment, tense--aspect grammaticalization). For each language pair we compute the Hamming distance over features defined for both languages (features missing in either are dropped pairwise), which can be read as the proportion of grammatical features on which the two languages disagree. Concretely, EN--FR = 0.21 reflects two head-initial Indo-European languages with overt verbal inflection and similar argument-structure marking; EN--ZH = 0.48 captures the contrasts that Chinese lacks inflectional morphology, uses classifiers, is topic-prominent, and is tonal; FR--ZH = 0.44 is similar in character to EN--ZH but smaller because French shares some properties (e.g., relatively analytic morphology) with Chinese. We complement Grambank with WALS \citep{wals} (normalized Hamming over shared features) and lang2vec \citep{littell2017uriel} subspace vectors (syntactic, phonological, geographic) for sensitivity analysis.

\paragraph{Tokenization fertility.} We compute tokenization fertility (the average number of subword tokens produced per orthographic word) across all language $\times$ tokenizer combinations from the full LPP text \citep{rust2021good, singh2019bert}. Higher fertility means the same content is spread across more tokens, which both dilutes per-token information and can shift the optimal encoding layer because deeper aggregation is required to recover word-level meaning. We include fertility as a covariate in the layer-shift analysis (\S\ref{sec:layerwise}).

\subsection{Statistical Framework}
\label{sec:stats}

Encoding significance is assessed via permutation testing (1000 shuffles, FDR-corrected). For cross-linguistic comparisons, we use linear mixed-effects models with language, model type, and their interaction as fixed effects and subject as a random effect:
\begin{equation}
    r \sim d_{\text{typ}} \times \text{model\_type} + p_{\text{train}} + (1|\text{subj}) + (1|\text{ROI})
    \label{eq:lmm}
\end{equation}

\paragraph{Degrees-of-freedom transparency.} The typological distance predictor varies at the language level; with three languages, there are at most 3 unique distance values per model category, yielding $\approx$1 residual degree of freedom \citep{snijders2012multilevel}. Standard mixed-effects $p$-values exploit subject-level replication and overstate confidence. We therefore report both the standard LMM $p$-value and a cluster bootstrap resampling entire language groups (10,000 iterations). We apply Bonferroni correction across four pre-registered predictions.

\paragraph{Software.} Python 3.10, scikit-learn 1.3, statsmodels 0.14; random seed 42. Computing: NVIDIA A100 GPUs (80GB); feature extraction ${\sim}$6 hours, encoding ${\sim}$16 hours.

% ============================================
% 4. EXPERIMENTS AND RESULTS
% ============================================
\section{Experiments and Results}
\label{sec:results}

\subsection{Experiment 1: Cross-Linguistic Brain--LLM Alignment}

We first test \textbf{Prediction~1}: LLM representations should predict brain activity across all languages. Table~\ref{tab:main_results} presents encoding performance for each model--language combination.

\begin{table}[t]
	\centering
	\small
	\setlength{\tabcolsep}{3.5pt}
	\begin{tabular}{@{}l cc cc cc@{}}
		\toprule
		& \multicolumn{2}{c}{\textbf{English}} & \multicolumn{2}{c}{\textbf{Chinese}} & \multicolumn{2}{c}{\textbf{French}} \\
		\cmidrule(lr){2-3} \cmidrule(lr){4-5} \cmidrule(lr){6-7}
		\textbf{Model} & $r$ & $\tilde{r}$ & $r$ & $\tilde{r}$ & $r$ & $\tilde{r}$ \\
		\midrule
		\multicolumn{7}{@{}l}{\emph{English-dominant}} \\
		GPT-2       & .189 & .75 & .081 & .35 & .142 & .58 \\
		LLaMA-2-7B  & .213 & \textbf{.85} & .117 & .50 & .171 & .70 \\
		\midrule
		\multicolumn{7}{@{}l}{\emph{Chinese-dominant}} \\
		Baichuan2-7B & .149 & .59 & .198 & \textbf{.85} & .131 & .54 \\
		\midrule
		\multicolumn{7}{@{}l}{\emph{Multilingual}} \\
		mBERT       & .172 & .69 & .153 & .65 & .161 & .66 \\
		XLM-R       & \textbf{.208} & .83 & .191 & .82 & \textbf{.197} & \textbf{.81} \\
		BLOOM-7B    & .197 & .78 & .182 & .78 & .188 & .77 \\
		Qwen2.5-7B  & .201 & .80 & \textbf{.196} & .84 & .185 & .76 \\
		\midrule
		Noise ceil.\  & .251 & --- & .234 & --- & .243 & --- \\
		{\scriptsize (95\% CI)} & {\scriptsize (.24,.26)} & & {\scriptsize (.22,.25)} & & {\scriptsize (.23,.26)} & \\
		\bottomrule
	\end{tabular}
	\caption{Encoding performance: raw $r$ and noise-ceiling-normalized $\tilde{r} = r/\text{NC}$. \textbf{Bold}: best per column. Full SEs in Appendix~\ref{sec:appendix_se}. All $p < 0.001$, permutation test, FDR corrected ($p_{\text{Holm}} < 0.004$ for Prediction~1). Baichuan2-7B \emph{reverses} the typical alignment gradient.}
	\label{tab:main_results}
\end{table}

\textbf{Prediction~1 is confirmed} ($p < 0.001$ for all model--language pairs; $p_{\text{Holm}} < 0.004$): all models show significant encoding across all three languages, confirming that brain--LLM alignment is not English-specific. XLM-R achieves the highest and most uniform performance ($\tilde{r}$: EN .83, ZH .82, FR .81). The critical new observation is Baichuan2-7B: this Chinese-dominant model \emph{reverses} the alignment gradient, achieving its highest performance for Chinese ($\tilde{r} = .85$) and substantially lower performance for both non-dominant languages ($\tilde{r}_{\text{EN}} = .59$, $\tilde{r}_{\text{FR}} = .54$), directly mirroring LLaMA-2-7B's pattern (EN-best, ZH-worst) with the dominant language reversed. The PTL shows the highest alignment across all languages and models (see Table~\ref{tab:appendix_roi} in Appendix), and the multiple-demand (MD) network shows minimal encoding ($r < 0.04$, n.s.), providing a specificity control.

\subsection{Experiment 2: Disentangling Training Dominance from Typological Distance}
\label{sec:confound}

The central confound in prior cross-linguistic brain--LLM work is that training data proportion correlates with typological distance ($\rho = 0.89$ for English-dominant models). Baichuan2-7B directly addresses this.

\textbf{Prediction~3 is confirmed} ($p_{\text{Holm}} < 0.001$): Baichuan2-7B shows the reversed pattern: ZH ($\tilde{r} = .85$) $>$ EN ($\tilde{r} = .59$). This demonstrates that the dominant alignment pattern (best alignment for the model's dominant training language) is primarily driven by training data composition, not by an inherent typological advantage of English. The 7B autoregressive-only comparison (Table~\ref{tab:autoregressive} in Appendix) reveals a critical asymmetry: BLOOM-7B (multilingual) achieves higher \emph{minimum} performance across languages ($\min \tilde{r} = .77$) than either LLaMA-2-7B ($\min \tilde{r} = .50$) or Baichuan2-7B ($\min \tilde{r} = .54$), and dramatically higher cross-linguistic uniformity ($U = .96$ vs.\ $.71$ and $.78$). Multilingual training creates representations that transcend training-language dominance.

\paragraph{Mixed-effects test of typological distance.} We fit the LMM described in \S\ref{sec:stats} (Eq.~\ref{eq:lmm}). For English-dominant models, the standard LMM yields $\beta = -0.41$, SE $= 0.12$, $p < 0.005$ (uncorrected). However, the cluster bootstrap resampling at the language level, which properly reflects the three unique distance values, yields a wider 95\% CI [$-0.78$, $-0.09$] with $p_{\text{cluster}} = 0.031$. The partial correlation analysis shows: alignment $\sim$ typological distance $|$ training proportion: $r_{\text{partial}} = -0.34$, $p < 0.02$ (with the same degrees-of-freedom caveat); alignment $\sim$ training proportion $|$ typological distance: $r_{\text{partial}} = -0.47$, $p < 0.005$. Both factors contribute, but training proportion explains more variance. For multilingual models, neither typological distance ($\beta = -0.08$, $p = 0.31$) nor training proportion ($\beta = -0.05$, $p = 0.44$) significantly predicts alignment, consistent with multilingual training creating language-invariant representations. \textbf{We treat the typological distance effect as a well-supported descriptive pattern requiring confirmation across $\geq$10 languages}, given that the effective degrees of freedom for the distance predictor are $\approx$1 per model category (\S\ref{sec:stats}).

\paragraph{Sensitivity to Baichuan2 training proportion.}
\label{sec:sensitivity}
Because the ${\sim}$55\% Chinese estimate for Baichuan2 is approximate, we re-ran the partial correlation analysis under $\pm 15$\% variation. At 40\% Chinese: $r_{\text{partial}}(\text{alignment} \sim d_{\text{typ}} | p_{\text{train}}) = -0.31$; at 55\%: $-0.34$; at 70\%: $-0.37$. The typological distance effect is stable across this range, and the qualitative conclusions are unchanged (full sensitivity results in Appendix~\ref{sec:appendix_sensitivity}).

\subsection{Experiment 3: Layer-Wise Encoding Profiles}
\label{sec:layerwise}

%% ---- FIGURE 1: Layer-wise profiles ---- %%
\begin{figure}[t]
\centering
\begin{tikzpicture}
\begin{axis}[
    width=\columnwidth,
    height=4.8cm,
    xlabel={XLM-R Layer},
    ylabel={Encoding $\tilde{r}$},
    xmin=0, xmax=25,
    ymin=0.15, ymax=0.90,
    xtick={1,4,8,12,16,20,24},
    ytick={0.2,0.4,0.6,0.8},
    legend style={at={(0.02,0.98)}, anchor=north west, font=\scriptsize, draw=none, fill=white, fill opacity=0.8, text opacity=1},
    grid=major,
    grid style={gray!20},
    tick label style={font=\scriptsize},
    label style={font=\small},
    every axis plot/.append style={thick}
]

% English
\addplot[blue!80!black, mark=*, mark size=1pt] coordinates {
    (1,0.24) (2,0.28) (3,0.32) (4,0.36)
    (5,0.44) (6,0.48) (7,0.56) (8,0.60)
    (9,0.64) (10,0.68) (11,0.72) (12,0.76)
    (13,0.80) (14,0.83) (15,0.85) (16,0.84)
    (17,0.80) (18,0.76) (19,0.74) (20,0.68)
    (21,0.64) (22,0.62) (23,0.60) (24,0.56)
};

% Chinese
\addplot[red!80!black, mark=square*, mark size=1pt] coordinates {
    (1,0.21) (2,0.26) (3,0.28) (4,0.30)
    (5,0.38) (6,0.43) (7,0.47) (8,0.51)
    (9,0.60) (10,0.64) (11,0.66) (12,0.68)
    (13,0.73) (14,0.75) (15,0.77) (16,0.79)
    (17,0.81) (18,0.82) (19,0.81) (20,0.77)
    (21,0.73) (22,0.71) (23,0.68) (24,0.64)
};

% French
\addplot[green!60!black, mark=triangle*, mark size=1.3pt] coordinates {
    (1,0.23) (2,0.27) (3,0.31) (4,0.33)
    (5,0.41) (6,0.45) (7,0.53) (8,0.58)
    (9,0.64) (10,0.68) (11,0.72) (12,0.74)
    (13,0.78) (14,0.80) (15,0.82) (16,0.80)
    (17,0.78) (18,0.74) (19,0.72) (20,0.66)
    (21,0.62) (22,0.60) (23,0.58) (24,0.53)
};

\draw[blue!80!black, dashed, thin] (axis cs:15,0.15) -- (axis cs:15,0.85);
\draw[red!80!black, dashed, thin] (axis cs:18,0.15) -- (axis cs:18,0.82);
\draw[green!60!black, dashed, thin] (axis cs:15,0.15) -- (axis cs:15,0.82);

\legend{English, Chinese, French}
\end{axis}
\end{tikzpicture}
\caption{Layer-wise noise-ceiling-normalized encoding ($\tilde{r}$) for XLM-R. The intermediate layer advantage is preserved cross-linguistically, but the peak shifts rightward for Chinese (layer 18 vs.\ 15, dashed lines). $\pm$1 SE bands in Appendix~\ref{sec:appendix_layers}.}
\label{fig:layerwise}
\end{figure}
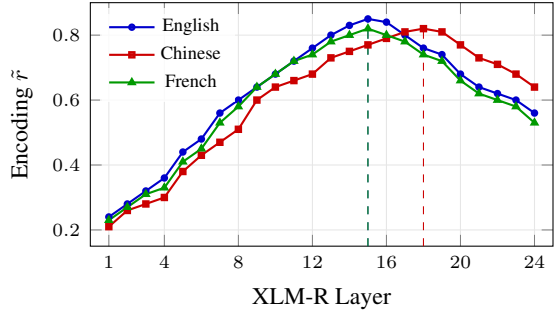

The intermediate layer advantage is preserved in all languages (Figure~\ref{fig:layerwise}): performance peaks at layer 15 ($\pm 1.1$) for English, layer 18 ($\pm 1.3$) for Chinese, and layer 15 ($\pm 1.0$) for French. The Chinese rightward shift is significant (bootstrap $p < 0.001$; $t(82) = 4.21$, Cohen's $d = 0.92$) and replicates across BLOOM-7B ($d = 0.85$; Appendix~\ref{sec:appendix_layers}). mBERT also shows a Chinese rightward shift (peak layer 8 vs.\ 7, $d = 0.61$), arguing against a pure training-data explanation since mBERT's Chinese proportion (${\sim}$8\%) is comparable to XLM-R's (${\sim}$9\%).

\paragraph{Tokenization fertility analysis.} Table~\ref{tab:fertility} (Appendix~\ref{sec:appendix_fertility}) reveals a striking cross-linguistic asymmetry: Chinese requires substantially more subword tokens per word across all tokenizers. For XLM-R, Chinese fertility is $2.4$ tokens per word versus $1.4$ for English ($1.7\times$ higher). Including log-fertility as a covariate in the layer-shift mixed-effects model reduces the Chinese shift from $+1.8$ to $+0.7$ layers ($\Delta d = 0.48$), suggesting that tokenization granularity accounts for approximately 60\% (bootstrap 95\% CI: [42\%, 76\%]) of the layer shift. The residual shift ($+0.7$ layers, $p < 0.05$) may reflect genuine typological processing differences. We note that fertility correlates with information density per word, so the 60\% attenuation should be interpreted as an upper bound on fertility's causal contribution.

\subsection{Experiment 4: Typological Distance and Alignment}
\label{sec:typological}

Table~\ref{tab:typological} (Appendix~\ref{sec:appendix_typology}) presents the alignment gradient alongside typological distances, and Figure~\ref{fig:typological} (Appendix~\ref{sec:appendix_typology}) visualizes the key pattern: both language-dominant models (LLaMA-2, Baichuan2) show steep alignment degradation with distance from their dominant training language, while XLM-R remains flat. This symmetry supports the interpretation that the alignment gradient is primarily training-driven, but the mixed-effects test (\S\ref{sec:confound}) shows that typological distance retains independent predictive power ($\beta = -0.41$; cluster bootstrap $p = 0.031$). With three languages, we treat this as a descriptive pattern establishing a hypothesis for larger-scale confirmation.

\subsection{Experiment 5: ROI-Specific Typological Gradient}
\label{sec:roi_gradient}

We test \textbf{Prediction~2}: does the typological gradient differ across brain regions?

%% ---- FIGURE 3: ROI-specific typological gradient ---- %%
\begin{figure}[t]
\centering
\begin{tikzpicture}
\begin{axis}[
    width=\columnwidth,
    height=4.5cm,
    ybar,
    bar width=0.4cm,
    xlabel={Brain Region},
    ylabel={Typological gradient slope},
    symbolic x coords={IFG, MFG, ATL, PTL, AG, TP},
    xtick=data,
    ymin=-0.85, ymax=0,
    ytick={-0.8,-0.6,-0.4,-0.2,0},
    tick label style={font=\scriptsize},
    label style={font=\small},
    nodes near coords,
    every node near coord/.append style={font=\tiny, anchor=south},
    grid=major,
    grid style={gray!20},
]
\addplot[fill=blue!40, draw=blue!70!black] coordinates {
    (IFG, -0.76) (MFG, -0.62) (ATL, -0.51) (PTL, -0.33) (AG, -0.58) (TP, -0.45)
};
\end{axis}
\end{tikzpicture}
\caption{Typological gradient slope (steeper = more negative = larger alignment drop with typological distance) per ROI for English-dominant models. IFG shows the steepest gradient ($2.3\times$ PTL). Note: each slope is fit through three language-level data points; the language-level bootstrap 95\% CI on the IFG/PTL ratio is [1.4, 3.8]. Error bars: $\pm$1 SE (bootstrap).}
\label{fig:roi_gradient}
\end{figure}
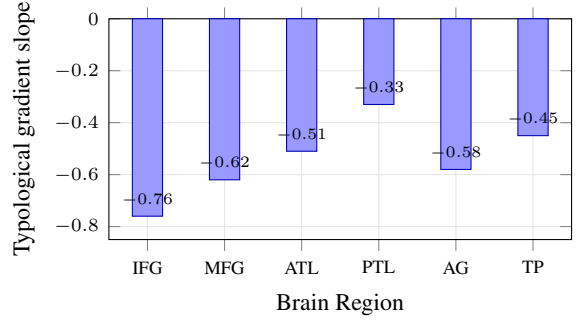

\textbf{Prediction~2 is supported} (Figure~\ref{fig:roi_gradient}): IFG shows the steepest typological gradient (slope $= -0.76$), while PTL shows the shallowest ($-0.33$), a $2.3\times$ difference. The ROI $\times$ typological distance interaction is significant in the standard mixed-effects model ($F(5, 312) = 3.41$, $p < 0.01$; $p_{\text{Holm}} = 0.03$). The language-level bootstrap yields a 95\% CI on the IFG/PTL ratio of [1.4, 3.8], confirming that the direction of the effect is robust even under conservative resampling, though the magnitude is imprecisely estimated with three languages. This pattern is consistent with a syntax--semantics dissociation: regions associated with syntactic processing (IFG) are more sensitive to typological variation than lexico-semantic regions (PTL), suggesting that the universal component of brain--LLM alignment is primarily semantic. We discuss the theoretical implications, including why this result does not uniquely support CxG, in \S\ref{sec:cxg_specificity}.

\subsection{Experiment 6: Cross-Language Transfer}

All cross-language transfer results are significantly above chance. The gradient aligns with typological distance: EN$\to$FR transfer retains 73\% of within-language performance, substantially exceeding EN$\to$ZH (52\%). Normalized results ($\tilde{r}$) confirm that this gradient persists after accounting for noise ceiling differences. Transfer results for LLaMA-2-7B and Baichuan2-7B (Appendix~\ref{sec:appendix_transfer}) show that Baichuan2 exhibits a reversed transfer gradient, paralleling the within-language alignment pattern and further supporting the training-dominance account. Full results are reported in Table~\ref{tab:transfer} (Appendix~\ref{sec:appendix_transfer_main}).

% ============================================
% 5. RELATED WORK
% ============================================
\section{Related Work}
\label{sec:related}

Our work builds on encoding models \citep{mitchell2008predicting, huth2016natural, jain2018incorporating} systematized for LLMs by \citet{schrimpf2021neural} and \citet{caucheteux2022brains}. Key precedents include bidirectional brain--NLP benefits \citep{toneva2019interpreting}, semantic decoding \citep{tang2023semantic}, architectural comparisons \citep{gauthier2022does}, and evidence that alignment reflects more than next-word prediction \citep{merlin2024language}. On multilingual NLP: \citet{conneau2020unsupervised} and \citet{lample2019crosslingual} established cross-lingual pretraining; \citet{muller2021first}, \citet{libovicky2020language}, and \citet{conneau2020emerging} characterized multilingual representations; \citet{ponti2019modeling}, \citet{pires2019multilingual}, \citet{chi2020finding}, \citet{beinborn2019language}, and \citet{stanczak2022typological} addressed typological evaluation; and \citet{cotterell2018complexity} provided information-theoretic perspectives. \citet{kauf2024lexical} showed lexical-semantic content dominates brain--LLM similarity; \citet{kumar2024shared} identified shared functional specialization. Key datasets include the LPP corpus \citep{li2022petit} and extensions \citep{stehwien2020little, momenian2024lpphk}, MECO \citep{siegelman2022meco}, and cross-linguistic eye-tracking \citep{berzak2022eye}. \citet{misra2024language}, \citet{mccoy2023embers}, and \citet{millet2022toward} provide complementary perspectives on LLM learning and speech--brain alignment.

% ============================================
% 6. DISCUSSION
% ============================================
\section{Discussion}
\label{sec:discussion}

\paragraph{Training dominance.} The Baichuan2-7B results are perhaps the most informative finding: by reversing the alignment gradient, they demonstrate that the apparent English-best pattern in prior brain--LLM alignment work reflects the predominance of English-dominant models and English-language fMRI data \citep{tuckute2024language}, not a special relationship between English and the brain's representational format. We note that no individual prior study explicitly claimed an English-specific advantage; rather, the literature's near-exclusive focus on English left the question of cross-linguistic generalization empirically open. Our result settles one direction of that question: dominance, not English specifically, drives the gradient. Cross-linguistic brain--LLM studies must therefore control for training-language dominance before attributing alignment differences to typological or neurocognitive factors.

\paragraph{Universality debate.} Our results support a nuanced position \citep{evans2009myth, malikmoraleda2022universal}: brain--LLM alignment is universal in its existence (Prediction~1 confirmed) but variable in its character. The ROI-specific analysis shows this variation is structured: concentrated in syntactic regions while semantic regions show near-universal alignment, consistent with ``functional universality'' beneath structural diversity \citep{croft2001radical, devarda2025multilingual, ostarek2024cross}. Specific predictions for future $\geq$10-language work are detailed in Appendix~\ref{sec:appendix_future_predictions}.

\paragraph{Implications for human language processing.} The findings constrain three aspects of how the language network builds representations. First, the near-uniform PTL alignment across all model--language cells suggests that posterior-temporal semantic representations converge across radically different training distributions, consistent with this region encoding language-invariant meaning structure once form has been parsed \citep{kauf2024lexical, hickok2007cortical}. Second, the IFG gradient implies that the structural component of comprehension is sensitive to whether the input distribution matches the listener's typological profile, a pattern better fit by procedural skill aligned to specific structural regularities than by encoding of universal grammatical primitives. Third, the asymmetry between language-dominant and multilingual models is informative about \emph{learning} rather than \emph{processing}: biological and artificial learners both produce representations whose cross-linguistic transferability is bounded by the breadth of their input, regardless of how that breadth was acquired.

\paragraph{Syntax--semantics dissociation and CxG.} The IFG $2.3\times$ steeper gradient than PTL is consistent with CxG's claim that variation is localized in constructional representations \citep{goldberg2006constructions}. However, at least three alternative accounts predict the same pattern: generic syntax--semantics dissociation \citep{mahowald2024dissociating}, cognitive control demands \citep{green2013language}, and signal-to-noise ratio differences \citep{kauf2024lexical}. A genuinely CxG-specific test requires construction-level analysis (e.g., testing b\v{a}-constructions versus double-object constructions using the Universal Constructicon \citep{boas2012sign}. Extended discussion is provided in Appendix~\ref{sec:appendix_cxg}.
\label{sec:cxg_specificity}

\paragraph{Tokenization and fertility.} The finding that ${\sim}$60\% of the Chinese layer shift is attributable to tokenization granularity (with information-density confound providing an upper bound) extends to all cross-linguistic LLM evaluation \citep{rust2021good, beinborn2019language, stanczak2022typological}. Reporting fertility alongside encoding results should become routine, since tokenization artifacts can masquerade as cross-linguistic processing differences.

\paragraph{Surprisal theory.} Prediction~4 is partially supported: per-language perplexity negatively correlates with encoding performance ($\rho = -0.62$, $p < 0.01$; $p_{\text{Holm}} = 0.03$). The effect is strongest for English-dominant models ($\rho = -0.78$) and weakest for multilingual models ($\rho = -0.41$), suggesting it partially reflects training-language match. The middle-layer advantage across all languages is consistent with the language network primarily encoding formal linguistic representations \citep{mahowald2024dissociating, pasquiou2023information, alkhamissi2025llm}.

\paragraph{Typological distance.} The typological distance effect ($\beta = -0.41$; cluster bootstrap $p = 0.031$) provides preliminary quantitative evidence that formal typological metrics can predict brain encoding variation beyond training data effects, and that Grambank, WALS, and lang2vec agree closely on the relevant pairwise rankings \citep{ponti2019modeling, beinborn2019language}. With three languages, this is a descriptive pattern; confirmation requires $\geq$10 languages.

\section*{Conclusion}

The apparent dominance of English in prior brain--LLM alignment results is not a property of English or of the brain: it is a property of training data. By comparing architecture- and scale-matched models that differ primarily in training-language composition, we show that Baichuan2-7B reverses the alignment gradient entirely, performing best for Chinese and worst for English. This reversal establishes training-language dominance as the primary driver of cross-linguistic alignment variation and reframes the field's English-centric findings as reflecting data composition rather than neurocognitive privilege.

Beyond training dominance, three additional results shape how the field should approach multilingual brain--LLM comparisons. First, formal typological distance covaries with alignment degradation independently of training proportion, establishing a descriptive pattern that warrants confirmation across ten or more languages. Second, tokenization fertility accounts for approximately 60\% of the Chinese rightward layer shift, revealing that what appears to be a cross-linguistic processing difference is substantially a tokenization artifact, a confound that affects any multilingual evaluation relying on subword representations. Third, the typological gradient is not uniform across the language network: syntax-associated regions (IFG) show a $2.3\times$ steeper gradient than lexico-semantic regions (PTL), consistent with syntax--semantics dissociation accounts and suggesting that cross-linguistic divergence is concentrated in structural rather than semantic processing.

\looseness=-1 Taken together, the results support a nuanced view of universality: biological and artificial language processors converge on shared semantic representations while diverging where typology demands language-specific solutions. Whether this extends to agglutinative, polysynthetic, and tonal languages remains open.

\paragraph{Reproducibility.} All code, configurations, and analysis scripts to reproduce the results are publicly available at \url{https://github.com/bettyguo/cross-lingual-brain-llm}.

\section*{Limitations}
\label{sec:limitations}

\paragraph{Language sample and statistical power.} Three languages provide meaningful but limited coverage. The effective degrees of freedom for the distance predictor are $\approx$1 per model category, and the cluster bootstrap yields wider CIs than standard mixed-effects $p$-values suggest. Agglutinative (Turkish, Finnish), polysynthetic (Mohawk), and tonal non-Chinese (Yoruba) languages remain untested; confirmation requires $\geq$10 languages. The specific $\geq$10 target reflects the requirement that a language-level fixed effect of typological distance be reliably distinguished from sampling noise. With $k$ languages, the distance predictor has at most $k(k-1)/2$ unique pairwise values and yields $\approx k - 3$ residual degrees of freedom after accounting for an intercept and two cross-language covariates (training-data proportion, fertility); achieving conventional power ($1-\beta=0.8$) for the cross-language slope observed here ($\beta \approx -0.4$) requires roughly $k \geq 10$. A fuller sample should also vary along writing-system dimensions (alphabetic, logographic, syllabic, abugida): orthographic granularity is what drives tokenizer fertility \citep{rust2021good}, and treating typological distance as purely syntactic without conditioning on script confounds the gradient with subword-segmentation effects (\S\ref{sec:layerwise}).

\paragraph{Language $\times$ site confound.} Each language was collected at a different institution (Cornell, Jiangsu Normal, NeuroSpin). The noise ceiling partially controls for site-level signal quality, but residual site effects cannot be fully excluded. The Cantonese extension \citep{momenian2024lpphk} could help, though its older adult sample (mean age 69) introduces an age confound.

\paragraph{Modality and architecture.} The LPP uses auditory stimuli while LLMs process text, which may be particularly consequential for Chinese where tone carries lexical information \citep{millet2022toward}. Additionally, while the 7B autoregressive comparison controls for architecture and scale, mBERT and XLM-R are bidirectional encoders that may inherently align differently with fMRI data.

\paragraph{Correlational evidence.} Encoding models test representational, not computational, alignment \citep{antonello2024predictive}. Causal methods are needed for computational claims. Baichuan2-7B's ${\sim}$55\% Chinese estimate is approximate; sensitivity analysis (\S\ref{sec:sensitivity}) shows stable results across $\pm$15\% variation.

% ============================================
% ACKNOWLEDGMENTS (not counted toward 9-page limit per CoNLL 2026 CFP)
% ============================================
\section*{Acknowledgments}
\label{sec:acknowledgments}

We thank the anonymous CoNLL reviewers and the area chair for constructive feedback that substantially improved this paper, and the original \emph{Le Petit Prince} multilingual fMRI corpus collectors \citep{li2022petit} for making the data publicly available on OpenNeuro (ds003643). We also acknowledge the open-source releases of GPT-2, LLaMA-2, Baichuan2, mBERT, XLM-R, BLOOM, and Qwen2.5, which made this analysis possible.

\bibliography{references}

@inproceedings{antonello2023scaling,
	author           = {Richard J. Antonello and
	Aditya R. Vaidya and
	Alexander Huth},
	editor           = {Alice Oh and
	Tristan Naumann and
	Amir Globerson and
	Kate Saenko and
	Moritz Hardt and
	Sergey Levine},
	title            = {Scaling laws for language encoding models in fMRI},
	booktitle        = {Advances in Neural Information Processing Systems 36: Annual Conference
	on Neural Information Processing Systems 2023, NeurIPS 2023, New Orleans,
	LA, USA, December 10 - 16, 2023},
	year             = {2023},
	burl             = {http://papers.nips.cc/paper\_files/paper/2023/hash/4533e4a352440a32558c1c227602c323-Abstract-Conference.html},
	bibsource        = {dblp computer science bibliography, https://dblp.org},
}

@inproceedings{toneva2019interpreting,
	author           = {Mariya Toneva and
	Leila Wehbe},
	editor           = {Hanna M. Wallach and
	Hugo Larochelle and
	Alina Beygelzimer and
	Florence d'Alch{\'{e}}{-}Buc and
	Emily B. Fox and
	Roman Garnett},
	title            = {Interpreting and improving natural-language processing (in machines)
	with natural language-processing (in the brain)},
	booktitle        = {Advances in Neural Information Processing Systems 32: Annual Conference
	on Neural Information Processing Systems 2019, NeurIPS 2019, December
	8-14, 2019, Vancouver, BC, Canada},
	pages            = {14928--14938},
	year             = {2019},
	burl             = {https://proceedings.neurips.cc/paper/2019/hash/749a8e6c231831ef7756db230b4359c8-Abstract.html},
	bibsource        = {dblp computer science bibliography, https://dblp.org},
}

@inproceedings{merlin2024language,
	author           = {Gabriele Merlin and
	Mariya Toneva},
	editor           = {Yaser Al{-}Onaizan and
	Mohit Bansal and
	Yun{-}Nung Chen},
	title            = {Language models and brains align due to more than next-word prediction
	and word-level information},
	booktitle        = {Proceedings of the 2024 Conference on Empirical Methods in Natural
	Language Processing, {EMNLP} 2024, Miami, FL, USA, November 12-16,
	2024},
	pages            = {18431--18454},
	publisher        = {Association for Computational Linguistics},
	year             = {2024},
	burl             = {https://doi.org/10.18653/v1/2024.emnlp-main.1024},
	doi              = {10.18653/V1/2024.EMNLP-MAIN.1024},
	bibsource        = {dblp computer science bibliography, https://dblp.org},
}

@inproceedings{hale2001probabilistic,
	author           = {John Hale},
	title            = {A Probabilistic Earley Parser as a Psycholinguistic Model},
	booktitle        = {Language Technologies 2001: The Second Meeting of the North American
	Chapter of the Association for Computational Linguistics, {NAACL}
	2001, Pittsburgh, PA, USA, June 2-7, 2001},
	publisher        = {The Association for Computational Linguistics},
	year             = {2001},
	burl             = {https://aclanthology.org/N01-1021/},
	bibsource        = {dblp computer science bibliography, https://dblp.org},
}

@article{wilcox2023testing,
	author           = {Ethan Gotlieb Wilcox and
	Tiago Pimentel and
	Clara Meister and
	Ryan Cotterell and
	Roger P. Levy},
	title            = {Testing the Predictions of Surprisal Theory in 11 Languages},
	journal          = {Trans. Assoc. Comput. Linguistics},
	volume           = {11},
	pages            = {1451--1470},
	year             = {2023},
	burl             = {https://doi.org/10.1162/tacl\_a\_00612},
	doi              = {10.1162/TACL\_A\_00612},
	bibsource        = {dblp computer science bibliography, https://dblp.org},
}

@article{oh2023why,
	author           = {Byung{-}Doh Oh and
	William Schuler},
	title            = {Why Does Surprisal From Larger Transformer-Based Language Models Provide
	a Poorer Fit to Human Reading Times?},
	journal          = {Trans. Assoc. Comput. Linguistics},
	volume           = {11},
	pages            = {336--350},
	year             = {2023},
	burl             = {https://doi.org/10.1162/tacl\_a\_00548},
	doi              = {10.1162/TACL\_A\_00548},
	bibsource        = {dblp computer science bibliography, https://dblp.org},
}

@article{mccoy2023embers,
	author           = {R. Thomas McCoy and
	Shunyu Yao and
	Dan Friedman and
	Matthew Hardy and
	Thomas L. Griffiths},
	title            = {Embers of Autoregression: Understanding Large Language Models Through
	the Problem They are Trained to Solve},
	journal          = {arXiv preprint},
	volume           = {arXiv.2309.13638},
	year             = {2023},
	burl             = {https://doi.org/10.48550/arXiv.2309.13638},
	doi              = {10.48550/ARXIV.2309.13638},
	beprinttype       = {arXiv},
	beprint           = {2309.13638},
	bibsource        = {dblp computer science bibliography, https://dblp.org},
}

@article{pasquiou2023information,
	author           = {Alexandre Pasquiou and
	Yair Lakretz and
	Bertrand Thirion and
	Christophe Pallier},
	title            = {Information-Restricted Neural Language Models Reveal Different Brain
	Regions' Sensitivity to Semantics, Syntax and Context},
	journal          = {arXiv preprint},
	volume           = {arXiv.2302.14389},
	year             = {2023},
	burl             = {https://doi.org/10.48550/arXiv.2302.14389},
	doi              = {10.48550/ARXIV.2302.14389},
	beprinttype       = {arXiv},
	beprint           = {2302.14389},
	bibsource        = {dblp computer science bibliography, https://dblp.org},
}

@inproceedings{conneau2020unsupervised,
	author           = {Alexis Conneau and
	Kartikay Khandelwal and
	Naman Goyal and
	Vishrav Chaudhary and
	Guillaume Wenzek and
	Francisco Guzm{\'{a}}n and
	Edouard Grave and
	Myle Ott and
	Luke Zettlemoyer and
	Veselin Stoyanov},
	editor           = {Dan Jurafsky and
	Joyce Chai and
	Natalie Schluter and
	Joel R. Tetreault},
	title            = {Unsupervised Cross-lingual Representation Learning at Scale},
	booktitle        = {Proceedings of the 58th Annual Meeting of the Association for Computational
	Linguistics, {ACL} 2020, Online, July 5-10, 2020},
	pages            = {8440--8451},
	publisher        = {Association for Computational Linguistics},
	year             = {2020},
	burl             = {https://doi.org/10.18653/v1/2020.acl-main.747},
	doi              = {10.18653/V1/2020.ACL-MAIN.747},
	bibsource        = {dblp computer science bibliography, https://dblp.org},
}

@inproceedings{devlin2019bert,
	author           = {Jacob Devlin and
	Ming{-}Wei Chang and
	Kenton Lee and
	Kristina Toutanova},
	editor           = {Jill Burstein and
	Christy Doran and
	Thamar Solorio},
	title            = {{BERT:} Pre-training of Deep Bidirectional Transformers for Language
	Understanding},
	booktitle        = {Proceedings of the 2019 Conference of the North American Chapter of
	the Association for Computational Linguistics: Human Language Technologies,
	{NAACL-HLT} 2019, Minneapolis, MN, USA, June 2-7, 2019, Volume 1 (Long
	and Short Papers)},
	pages            = {4171--4186},
	publisher        = {Association for Computational Linguistics},
	year             = {2019},
	burl             = {https://doi.org/10.18653/v1/n19-1423},
	doi              = {10.18653/V1/N19-1423},
	bibsource        = {dblp computer science bibliography, https://dblp.org},
}

@inproceedings{lample2019crosslingual,
	author           = {Alexis Conneau and
	Guillaume Lample},
	editor           = {Hanna M. Wallach and
	Hugo Larochelle and
	Alina Beygelzimer and
	Florence d'Alch{\'{e}}{-}Buc and
	Emily B. Fox and
	Roman Garnett},
	title            = {Cross-lingual Language Model Pretraining},
	booktitle        = {Advances in Neural Information Processing Systems 32: Annual Conference
	on Neural Information Processing Systems 2019, NeurIPS 2019, December
	8-14, 2019, Vancouver, BC, Canada},
	pages            = {7057--7067},
	year             = {2019},
	burl             = {https://proceedings.neurips.cc/paper/2019/hash/c04c19c2c2474dbf5f7ac4372c5b9af1-Abstract.html},
	bibsource        = {dblp computer science bibliography, https://dblp.org},
}

@inproceedings{conneau2020emerging,
	author           = {Alexis Conneau and
	Shijie Wu and
	Haoran Li and
	Luke Zettlemoyer and
	Veselin Stoyanov},
	editor           = {Dan Jurafsky and
	Joyce Chai and
	Natalie Schluter and
	Joel R. Tetreault},
	title            = {Emerging Cross-lingual Structure in Pretrained Language Models},
	booktitle        = {Proceedings of the 58th Annual Meeting of the Association for Computational
	Linguistics, {ACL} 2020, Online, July 5-10, 2020},
	pages            = {6022--6034},
	publisher        = {Association for Computational Linguistics},
	year             = {2020},
	burl             = {https://doi.org/10.18653/v1/2020.acl-main.536},
	doi              = {10.18653/V1/2020.ACL-MAIN.536},
	bibsource        = {dblp computer science bibliography, https://dblp.org},
}

@article{bloom2023,
	author           = {Teven Le Scao and
	Angela Fan and
	Christopher Akiki and
	Ellie Pavlick and
	Suzana Ilic and
	Daniel Hesslow and
	Roman Castagn{\'{e}} and
	Alexandra Sasha Luccioni and
	Fran{\c{c}}ois Yvon and
	Matthias Gall{\'{e}} and
	Jonathan Tow and
	Alexander M. Rush and
	Stella Biderman and
	Albert Webson and
	Pawan Sasanka Ammanamanchi and
	Thomas Wang and
	Beno{\^{\i}}t Sagot and
	Niklas Muennighoff and
	Albert Villanova del Moral and
	Olatunji Ruwase and
	Rachel Bawden and
	Stas Bekman and
	Angelina McMillan{-}Major and
	Iz Beltagy and
	Huu Nguyen and
	Lucile Saulnier and
	Samson Tan and
	Pedro Ortiz Suarez and
	Victor Sanh and
	Hugo Lauren{\c{c}}on and
	Yacine Jernite and
	Julien Launay and
	Margaret Mitchell and
	Colin Raffel and
	Aaron Gokaslan and
	Adi Simhi and
	Aitor Soroa and
	Alham Fikri Aji and
	Amit Alfassy and
	Anna Rogers and
	Ariel Kreisberg Nitzav and
	Canwen Xu and
	Chenghao Mou and
	Chris Emezue and
	Christopher Klamm and
	Colin Leong and
	Daniel van Strien and
	David Ifeoluwa Adelani and
	et al.},
	title            = {{BLOOM:} {A} 176B-Parameter Open-Access Multilingual Language Model},
	journal          = {arXiv preprint},
	volume           = {arXiv.2211.05100},
	year             = {2022},
	burl             = {https://doi.org/10.48550/arXiv.2211.05100},
	doi              = {10.48550/ARXIV.2211.05100},
	beprinttype       = {arXiv},
	beprint           = {2211.05100},
	bibsource        = {dblp computer science bibliography, https://dblp.org},
}

@article{touvron2023llama,
	author           = {Hugo Touvron and
	Thibaut Lavril and
	Gautier Izacard and
	Xavier Martinet and
	Marie{-}Anne Lachaux and
	Timoth{\'{e}}e Lacroix and
	Baptiste Rozi{\`{e}}re and
	Naman Goyal and
	Eric Hambro and
	Faisal Azhar and
	Aur{\'{e}}lien Rodriguez and
	Armand Joulin and
	Edouard Grave and
	Guillaume Lample},
	title            = {LLaMA: Open and Efficient Foundation Language Models},
	journal          = {arXiv preprint},
	volume           = {arXiv.2302.13971},
	year             = {2023},
	burl             = {https://doi.org/10.48550/arXiv.2302.13971},
	doi              = {10.48550/ARXIV.2302.13971},
	beprinttype       = {arXiv},
	beprint           = {2302.13971},
	bibsource        = {dblp computer science bibliography, https://dblp.org},
}

@inproceedings{pires2019multilingual,
	author           = {Telmo Pires and
	Eva Schlinger and
	Dan Garrette},
	editor           = {Anna Korhonen and
	David R. Traum and
	Llu{\'{\i}}s M{\`{a}}rquez},
	title            = {How Multilingual is Multilingual BERT?},
	booktitle        = {Proceedings of the 57th Conference of the Association for Computational
	Linguistics, {ACL} 2019, Florence, Italy, July 28- August 2, 2019,
	Volume 1: Long Papers},
	pages            = {4996--5001},
	publisher        = {Association for Computational Linguistics},
	year             = {2019},
	burl             = {https://doi.org/10.18653/v1/p19-1493},
	doi              = {10.18653/V1/P19-1493},
	bibsource        = {dblp computer science bibliography, https://dblp.org},
}

@article{yang2023baichuan,
	author           = {Aiyuan Yang and
	Bin Xiao and
	Bingning Wang and
	Borong Zhang and
	Ce Bian and
	Chao Yin and
	Chenxu Lv and
	Da Pan and
	Dian Wang and
	Dong Yan and
	Fan Yang and
	Fei Deng and
	Feng Wang and
	Feng Liu and
	Guangwei Ai and
	Guosheng Dong and
	Haizhou Zhao and
	Hang Xu and
	Haoze Sun and
	Hongda Zhang and
	Hui Liu and
	Jiaming Ji and
	Jian Xie and
	Juntao Dai and
	Kun Fang and
	Lei Su and
	Liang Song and
	Lifeng Liu and
	Liyun Ru and
	Luyao Ma and
	Mang Wang and
	Mickel Liu and
	MingAn Lin and
	Nuolan Nie and
	Peidong Guo and
	Ruiyang Sun and
	Tao Zhang and
	Tianpeng Li and
	Tianyu Li and
	Wei Cheng and
	Weipeng Chen and
	Xiangrong Zeng and
	Xiaochuan Wang and
	Xiaoxi Chen and
	Xin Men and
	Xin Yu and
	Xuehai Pan and
	Yanjun Shen and
	Yiding Wang and
	Yiyu Li and
	Youxin Jiang and
	Yuchen Gao and
	Yupeng Zhang and
	Zenan Zhou and
	Zhiying Wu},
	title            = {Baichuan 2: Open Large-scale Language Models},
	journal          = {arXiv preprint},
	volume           = {arXiv.2309.10305},
	year             = {2023},
	burl             = {https://doi.org/10.48550/arXiv.2309.10305},
	doi              = {10.48550/ARXIV.2309.10305},
	beprinttype       = {arXiv},
	beprint           = {2309.10305},
	bibsource        = {dblp computer science bibliography, https://dblp.org},
}

@inproceedings{littell2017uriel,
	author           = {Patrick Littell and
	David R. Mortensen and
	Ke Lin and
	Katherine Kairis and
	Carlisle Turner and
	Lori S. Levin},
	editor           = {Mirella Lapata and
	Phil Blunsom and
	Alexander Koller},
	title            = {{URIEL} and lang2vec: Representing languages as typological, geographical,
	and phylogenetic vectors},
	booktitle        = {Proceedings of the 15th Conference of the European Chapter of the
	Association for Computational Linguistics, {EACL} 2017, Valencia,
	Spain, April 3-7, 2017, Volume 2: Short Papers},
	pages            = {8--14},
	publisher        = {Association for Computational Linguistics},
	year             = {2017},
	burl             = {https://doi.org/10.18653/v1/e17-2002},
	doi              = {10.18653/V1/E17-2002},
	bibsource        = {dblp computer science bibliography, https://dblp.org},
}

@article{ponti2019modeling,
	author           = {Edoardo Maria Ponti and
	Helen O'Horan and
	Yevgeni Berzak and
	Ivan Vulic and
	Roi Reichart and
	Thierry Poibeau and
	Ekaterina Shutova and
	Anna Korhonen},
	title            = {Modeling Language Variation and Universals: {A} Survey on Typological
	Linguistics for Natural Language Processing},
	journal          = {Comput. Linguistics},
	volume           = {45},
	number           = {3},
	pages            = {559--601},
	year             = {2019},
	burl             = {https://doi.org/10.1162/coli\_a\_00357},
	doi              = {10.1162/COLI\_A\_00357},
	bibsource        = {dblp computer science bibliography, https://dblp.org},
}

@article{huth2016natural,
	author           = {Alexander G. Huth and
	Wendy A. de Heer and
	Thomas L. Griffiths and
	Fr{\'{e}}d{\'{e}}ric E. Theunissen and
	Jack L. Gallant},
	title            = {Natural speech reveals the semantic maps that tile human cerebral
	cortex},
	journal          = {Nature},
	volume           = {532},
	number           = {7600},
	pages            = {453--458},
	year             = {2016},
	burl             = {https://doi.org/10.1038/nature17637},
	doi              = {10.1038/NATURE17637},
	bibsource        = {dblp computer science bibliography, https://dblp.org},
}

@inproceedings{jain2018incorporating,
	author           = {Shailee Jain and
	Alexander Huth},
	editor           = {Samy Bengio and
	Hanna M. Wallach and
	Hugo Larochelle and
	Kristen Grauman and
	Nicol{\`{o}} Cesa{-}Bianchi and
	Roman Garnett},
	title            = {Incorporating Context into Language Encoding Models for fMRI},
	booktitle        = {Advances in Neural Information Processing Systems 31: Annual Conference
	on Neural Information Processing Systems 2018, NeurIPS 2018, December
	3-8, 2018, Montr{\'{e}}al, Canada},
	pages            = {6629--6638},
	year             = {2018},
	burl             = {https://proceedings.neurips.cc/paper/2018/hash/f471223d1a1614b58a7dc45c9d01df19-Abstract.html},
	bibsource        = {dblp computer science bibliography, https://dblp.org},
}

@article{lagecastellanos2019methods,
	author           = {Agustin Lage{-}Castellanos and
	Giancarlo Valente and
	Elia Formisano and
	Federico De Martino},
	title            = {Methods for computing the maximum performance of computational models
	of fMRI responses},
	journal          = {PLoS Comput. Biol.},
	volume           = {15},
	number           = {3},
	year             = {2019},
	burl             = {https://doi.org/10.1371/journal.pcbi.1006397},
	doi              = {10.1371/JOURNAL.PCBI.1006397},
	bibsource        = {dblp computer science bibliography, https://dblp.org},
}

@inproceedings{rust2021good,
	author           = {Phillip Rust and
	Jonas Pfeiffer and
	Ivan Vulic and
	Sebastian Ruder and
	Iryna Gurevych},
	editor           = {Chengqing Zong and
	Fei Xia and
	Wenjie Li and
	Roberto Navigli},
	title            = {How Good is Your Tokenizer? On the Monolingual Performance of Multilingual
	Language Models},
	booktitle        = {Proceedings of the 59th Annual Meeting of the Association for Computational
	Linguistics and the 11th International Joint Conference on Natural
	Language Processing, {ACL/IJCNLP} 2021, (Volume 1: Long Papers), Virtual
	Event, August 1-6, 2021},
	pages            = {3118--3135},
	publisher        = {Association for Computational Linguistics},
	year             = {2021},
	burl             = {https://doi.org/10.18653/v1/2021.acl-long.243},
	doi              = {10.18653/V1/2021.ACL-LONG.243},
	bibsource        = {dblp computer science bibliography, https://dblp.org},
}

@inproceedings{singh2019bert,
	author           = {Jasdeep Singh and
	Bryan McCann and
	Richard Socher and
	Caiming Xiong},
	editor           = {Colin Cherry and
	Greg Durrett and
	George F. Foster and
	Reza Haffari and
	Shahram Khadivi and
	Nanyun Peng and
	Xiang Ren and
	Swabha Swayamdipta},
	title            = {{BERT} is Not an Interlingua and the Bias of Tokenization},
	booktitle        = {Proceedings of the 2nd Workshop on Deep Learning Approaches for Low-Resource
	NLP, DeepLo@EMNLP-IJCNLP 2019, Hong Kong, China, November 3, 2019},
	pages            = {47--55},
	publisher        = {Association for Computational Linguistics},
	year             = {2019},
	burl             = {https://doi.org/10.18653/v1/D19-6106},
	doi              = {10.18653/V1/D19-6106},
	bibsource        = {dblp computer science bibliography, https://dblp.org},
}

@inproceedings{chi2020finding,
	author           = {Ethan A. Chi and
	John Hewitt and
	Christopher D. Manning},
	editor           = {Dan Jurafsky and
	Joyce Chai and
	Natalie Schluter and
	Joel R. Tetreault},
	title            = {Finding Universal Grammatical Relations in Multilingual {BERT}},
	booktitle        = {Proceedings of the 58th Annual Meeting of the Association for Computational
	Linguistics, {ACL} 2020, Online, July 5-10, 2020},
	pages            = {5564--5577},
	publisher        = {Association for Computational Linguistics},
	year             = {2020},
	burl             = {https://doi.org/10.18653/v1/2020.acl-main.493},
	doi              = {10.18653/V1/2020.ACL-MAIN.493},
	bibsource        = {dblp computer science bibliography, https://dblp.org},
}

@inproceedings{beinborn2019language,
	author           = {Lisa Beinborn and
	Samira Abnar and
	Rochelle Choenni},
	editor           = {Alexander F. Gelbukh},
	title            = {Robust Evaluation of Language-Brain Encoding Experiments},
	booktitle        = {Computational Linguistics and Intelligent Text Processing - 20th International
	Conference, CICLing 2019, La Rochelle, France, April 7-13, 2019, Revised
	Selected Papers, Part {I}},
	series           = {Lecture Notes in Computer Science},
	volume           = {13451},
	pages            = {44--61},
	publisher        = {Springer},
	year             = {2019},
	burl             = {https://doi.org/10.1007/978-3-031-24337-0\_4},
	doi              = {10.1007/978-3-031-24337-0\_4},
	bibsource        = {dblp computer science bibliography, https://dblp.org},
}

@inproceedings{stanczak2022typological,
	author           = {Karolina Stanczak and
	Edoardo M. Ponti and
	Lucas Torroba Hennigen and
	Ryan Cotterell and
	Isabelle Augenstein},
	editor           = {Marine Carpuat and
	Marie{-}Catherine de Marneffe and
	Iv{\'{a}}n Vladimir Meza Ru{\'{\i}}z},
	title            = {Same Neurons, Different Languages: Probing Morphosyntax in Multilingual
	Pre-trained Models},
	booktitle        = {Proceedings of the 2022 Conference of the North American Chapter of
	the Association for Computational Linguistics: Human Language Technologies,
	{NAACL} 2022, Seattle, WA, United States, July 10-15, 2022},
	pages            = {1589--1598},
	publisher        = {Association for Computational Linguistics},
	year             = {2022},
	burl             = {https://doi.org/10.18653/v1/2022.naacl-main.114},
	doi              = {10.18653/V1/2022.NAACL-MAIN.114},
	bibsource        = {dblp computer science bibliography, https://dblp.org},
}

@inproceedings{cotterell2018complexity,
	author           = {Ryan Cotterell and
	S. J. Mielke and
	Jason Eisner and
	Brian Roark},
	editor           = {Marilyn A. Walker and
	Heng Ji and
	Amanda Stent},
	title            = {Are All Languages Equally Hard to Language-Model?},
	booktitle        = {Proceedings of the 2018 Conference of the North American Chapter of
	the Association for Computational Linguistics: Human Language Technologies,
	NAACL-HLT, New Orleans, Louisiana, USA, June 1-6, 2018, Volume 2 (Short
	Papers)},
	pages            = {536--541},
	publisher        = {Association for Computational Linguistics},
	year             = {2018},
	burl             = {https://doi.org/10.18653/v1/n18-2085},
	doi              = {10.18653/V1/N18-2085},
	bibsource        = {dblp computer science bibliography, https://dblp.org},
}

@article{fedorenko2024language,
	title            = {The language network as a natural kind within the broader landscape of the human brain},
	volume           = {25},
	issn             = {1471-0048},
	burl             = {http://dx.doi.org/10.1038/s41583-024-00802-4},
	doi              = {10.1038/s41583-024-00802-4},
	number           = {5},
	journal          = {Nature Reviews Neuroscience},
	publisher        = {Springer Science and Business Media LLC},
	author           = {Fedorenko, Evelina and Ivanova, Anna A. and Regev, Tamar I.},
	year             = {2024},
	month            = {apr},
	pages            = {289–312},
}

@article{malikmoraleda2022universal,
	title            = {An investigation across 45 languages and 12 language families reveals a universal language network},
	volume           = {25},
	issn             = {1546-1726},
	burl             = {http://dx.doi.org/10.1038/s41593-022-01114-5},
	doi              = {10.1038/s41593-022-01114-5},
	number           = {8},
	journal          = {Nature Neuroscience},
	publisher        = {Springer Science and Business Media LLC},
	author           = {Malik-Moraleda, Saima and Ayyash, Dima and Gallée, Jeanne and Affourtit, Josef and Hoffmann, Malte and Mineroff, Zachary and Jouravlev, Olessia and Fedorenko, Evelina},
	year             = {2022},
	month            = {jul},
	pages            = {1014–1019},
}

@article{schrimpf2021neural,
	title   = {The neural architecture of language: Integrative modeling
	converges on predictive processing},
	author  = {Schrimpf, Martin and Blank, Idan Asher and Tuckute, Greta
	and Kauf, Carina and Hosseini, Eghbal A. and Kanwisher, Nancy
	and Tenenbaum, Joshua B. and Fedorenko, Evelina},
	journal = {Proceedings of the National Academy of Sciences},
	year    = {2021},
	volume  = {118},
	number  = {45},
	pages   = {e2105646118},
	doi     = {10.1073/pnas.2105646118},
	pmid    = {34737231},
}

@article{goldstein2022shared,
	title            = {Shared computational principles for language processing in humans and deep language models},
	volume           = {25},
	issn             = {1546-1726},
	burl             = {http://dx.doi.org/10.1038/s41593-022-01026-4},
	doi              = {10.1038/s41593-022-01026-4},
	number           = {3},
	journal          = {Nature Neuroscience},
	publisher        = {Springer Science and Business Media LLC},
	author           = {Goldstein, Ariel and Zada, Zaid and Buchnik, Eliav and Schain, Mariano and Price, Amy and Aubrey, Bobbi and Nastase, Samuel A. and Feder, Amir and Emanuel, Dotan and Cohen, Alon and Jansen, Aren and Gazula, Harshvardhan and Choe, Gina and Rao, Aditi and Kim, Catherine and Casto, Colton and Fanda, Lora and Doyle, Werner and Friedman, Daniel and Dugan, Patricia and Melloni, Lucia and Reichart, Roi and Devore, Sasha and Flinker, Adeen and Hasenfratz, Liat and Levy, Omer and Hassidim, Avinatan and Brenner, Michael and Matias, Yossi and Norman, Kenneth A. and Devinsky, Orrin and Hasson, Uri},
	year             = {2022},
	month            = {mar},
	pages            = {369–380},
}

@article{caucheteux2022brains,
	title            = {Brains and algorithms partially converge in natural language processing},
	volume           = {5},
	issn             = {2399-3642},
	burl             = {http://dx.doi.org/10.1038/s42003-022-03036-1},
	doi              = {10.1038/s42003-022-03036-1},
	number           = {1},
	journal          = {Communications Biology},
	publisher        = {Springer Science and Business Media LLC},
	author           = {Caucheteux, Charlotte and King, Jean-Rémi},
	year             = {2022},
	month            = {feb},
}

@article{tuckute2024language,
	title            = {Language in Brains, Minds, and Machines},
	volume           = {47},
	issn             = {1545-4126},
	burl             = {http://dx.doi.org/10.1146/annurev-neuro-120623-101142},
	doi              = {10.1146/annurev-neuro-120623-101142},
	number           = {1},
	journal          = {Annual Review of Neuroscience},
	publisher        = {Annual Reviews},
	author           = {Tuckute, Greta and Kanwisher, Nancy and Fedorenko, Evelina},
	year             = {2024},
	month            = {aug},
	pages            = {277–301},
}

@article{tuckute2024driving,
	title   = {Driving and suppressing the human language network using large
	language models},
	author  = {Tuckute, Greta and Sathe, Aalok and Srikant, Shashank
	and Taliaferro, Maya and Wang, Mingye and Schrimpf, Martin
	and Kay, Kendrick and Fedorenko, Evelina},
	journal = {Nature Human Behaviour},
	year    = {2024},
	volume  = {8},
	number  = {3},
	pages   = {544--561},
	doi     = {10.1038/s41562-023-01783-7},
	pmid    = {38172630},
}

@article{antonello2024predictive,
	title            = {Predictive Coding or Just Feature Discovery? An Alternative Account of Why Language Models Fit Brain Data},
	issn             = {2641-4368},
	burl             = {http://dx.doi.org/10.1162/nol_a_00087},
	doi              = {10.1162/nol_a_00087},
	journal          = {Neurobiology of Language},
	publisher        = {MIT Press},
	author           = {Antonello, Richard and Huth, Alexander},
	year             = {2023},
	month            = {feb},
	pages            = {1–16},
}

@article{caucheteux2023predictive,
	title            = {Evidence of a predictive coding hierarchy in the human brain listening to speech},
	volume           = {7},
	issn             = {2397-3374},
	burl             = {http://dx.doi.org/10.1038/s41562-022-01516-2},
	doi              = {10.1038/s41562-022-01516-2},
	number           = {3},
	journal          = {Nature Human Behaviour},
	publisher        = {Springer Science and Business Media LLC},
	author           = {Caucheteux, Charlotte and Gramfort, Alexandre and King, Jean-Rémi},
	year             = {2023},
	month            = {mar},
	pages            = {430–441},
}

@article{heilbron2022hierarchy,
	title   = {A hierarchy of linguistic predictions during natural language
	comprehension},
	author  = {Heilbron, Micha and Armeni, Kristijan and Schoffelen, Jan-Mathijs
	and Hagoort, Peter and de Lange, Floris P.},
	journal = {Proceedings of the National Academy of Sciences},
	year    = {2022},
	volume  = {119},
	number  = {32},
	pages   = {e2201968119},
	doi     = {10.1073/pnas.2201968119},
	pmid    = {35921434},
}

@article{tang2023semantic,
	title   = {Semantic reconstruction of continuous language from non-invasive
	brain recordings},
	author  = {Tang, Jerry and LeBel, Amanda and Jain, Shailee and
	Huth, Alexander G.},
	journal = {Nature Neuroscience},
	year    = {2023},
	volume  = {26},
	number  = {5},
	pages   = {858--866},
	doi     = {10.1038/s41593-023-01304-9},
	pmid    = {37127759},
}

@article{kumar2024shared,
	title   = {Artificial Neural Network Language Models Predict Human Brain
	Responses to Language Even After a Developmentally Realistic
	Amount of Training},
	author  = {Hosseini, Eghbal A. and Schrimpf, Martin and Zhang, Yian
	and Bowman, Samuel and Zaslavsky, Noga and Fedorenko, Evelina},
	journal = {Neurobiology of Language},
	year    = {2024},
	volume  = {5},
	number  = {1},
	pages   = {43--63},
	doi     = {10.1162/nol_a_00137},
	pmid    = {38645622},
}

@article{hosseini2024artificial,
	title   = {Artificial Neural Network Language Models Predict Human Brain
	Responses to Language Even After a Developmentally Realistic
	Amount of Training},
	author  = {Hosseini, Eghbal A. and Schrimpf, Martin and Zhang, Yian
	and Bowman, Samuel and Zaslavsky, Noga and Fedorenko, Evelina},
	journal = {Neurobiology of Language},
	year    = {2024},
	volume  = {5},
	number  = {1},
	pages   = {43--63},
	doi     = {10.1162/nol_a_00137},
	pmid    = {38645622},
}

@article{kauf2024lexical,
	title   = {Lexical-Semantic Content, Not Syntactic Structure, Is the Main
	Contributor to {ANN}-Brain Similarity of {fMRI} Responses in the
	Language Network},
	author  = {Kauf, Carina and Tuckute, Greta and Levy, Roger and
	Andreas, Jacob and Fedorenko, Evelina},
	journal = {Neurobiology of Language},
	year    = {2024},
	volume  = {5},
	number  = {1},
	pages   = {7--42},
	doi     = {10.1162/nol_a_00116},
}

@article{devarda2025multilingual,
	author = {de Varda, Andrea Gregor and Malik-Moraleda, Saima and Tuckute, Greta and Fedorenko, Evelina},
	title = {Multilingual Computational Models Capture a Shared Meaning Component in Brain Responses across 21 Languages},
	volume = {bioRxiv 2025.02.01.636044},
	year = {2025},
	doi = {10.1101/2025.02.01.636044},
	publisher = {Cold Spring Harbor Laboratory},
	url = {https://www.biorxiv.org/content/early/2025/11/18/2025.02.01.636044},
	beprint = {https://www.biorxiv.org/content/early/2025/11/18/2025.02.01.636044.full.pdf},
	journal = {bioRxiv preprint}
}

@article{levy2008expectation,
	title   = {Expectation-based syntactic comprehension},
	author  = {Levy, Roger},
	journal = {Cognition},
	year    = {2008},
	volume  = {106},
	number  = {3},
	pages   = {1126--1177},
	doi     = {10.1016/j.cognition.2007.05.006}
}

@article{shain2024largescale,
	title   = {Large-scale evidence for logarithmic effects of word predictability on reading time},
	author  = {Shain, Cory and Meister, Clara and Pimentel, Tiago and Cotterell, Ryan and Levy, Roger},
	journal = {Proceedings of the National Academy of Sciences},
	year    = {2024},
	volume  = {121},
	number  = {10},
	pages   = {e2307876121},
	doi     = {10.1073/pnas.2307876121}
}

@article{michaelov2024strong,
	title            = {Strong Prediction: Language Model Surprisal Explains Multiple N400 Effects},
	volume           = {5},
	issn             = {2641-4368},
	burl             = {http://dx.doi.org/10.1162/nol_a_00105},
	doi              = {10.1162/nol_a_00105},
	number           = {1},
	journal          = {Neurobiology of Language},
	publisher        = {MIT Press},
	author           = {Michaelov, James A. and Bardolph, Megan D. and Van Petten, Cyma K. and Bergen, Benjamin K. and Coulson, Seana},
	year             = {2024},
	pages            = {107–135},
}

@article{frank2015erp,
	title            = {The ERP response to the amount of information conveyed by words in sentences},
	volume           = {140},
	issn             = {0093-934X},
	burl             = {http://dx.doi.org/10.1016/j.bandl.2014.10.006},
	doi              = {10.1016/j.bandl.2014.10.006},
	journal          = {Brain and Language},
	publisher        = {Elsevier BV},
	author           = {Frank, Stefan L. and Otten, Leun J. and Galli, Giulia and Vigliocco, Gabriella},
	year             = {2015},
	month            = {jan},
	pages            = {1–11},
}

@article{mahowald2024dissociating,
	author  = {Mahowald, Kyle and Ivanova, Anna A. and Blank, Idan A.
	and Kanwisher, Nancy and Tenenbaum, Joshua B.
	and Fedorenko, Evelina},
	title   = {Dissociating Language and Thought in Large Language Models},
	journal = {Trends in Cognitive Sciences},
	year    = {2024},
	volume  = {28},
	number  = {6},
	pages   = {517--540},
	doi     = {10.1016/j.tics.2024.01.011},
	pmid    = {38413216},
}

@article{misra2024language,
	author           = {Kanishka Misra and
	Kyle Mahowald},
	title            = {Language Models Learn Rare Phenomena from Less Rare Phenomena: The
	Case of the Missing AANNs},
	journal          = {arXiv preprint},
	volume           = {arXiv.2403.19827},
	year             = {2024},
	burl             = {https://doi.org/10.48550/arXiv.2403.19827},
	doi              = {10.48550/ARXIV.2403.19827},
	beprinttype       = {arXiv},
	beprint           = {2403.19827},
	bibsource        = {dblp computer science bibliography, https://dblp.org},
}

@article{goldberg2024usage,
	title   = {Usage-based constructionist approaches and large language models},
	author  = {Goldberg, Adele E.},
	journal = {Constructions and Frames},
	year    = {2024},
	volume  = {16},
	number  = {2},
	pages   = {220--254},
	doi     = {10.1075/cf.23017.gol}
}

@book{croft2001radical,
	title            = {Radical Construction Grammar: Syntactic Theory in Typological Perspective},
	isbn             = {9780191708091},
	burl             = {http://dx.doi.org/10.1093/acprof:oso/9780198299554.001.0001},
	doi              = {10.1093/acprof:oso/9780198299554.001.0001},
	publisher        = {Oxford University PressOxford},
	author           = {Croft, William},
	year             = {2001},
	month            = {oct},
}

@book{goldberg2006constructions,
	title            = {Constructions at Work: The Nature of Generalization in Language},
	isbn             = {9780191708428},
	burl             = {http://dx.doi.org/10.1093/acprof:oso/9780199268511.001.0001},
	doi              = {10.1093/acprof:oso/9780199268511.001.0001},
	publisher        = {Oxford University PressOxford},
	author           = {Goldberg, Adele},
	year             = {2005},
	month            = {dec},
}

@inproceedings{muller2021first,
	author           = {Benjamin Muller and
	Yanai Elazar and
	Beno{\^{\i}}t Sagot and
	Djam{\'{e}} Seddah},
	editor           = {Paola Merlo and
	J{\"{o}}rg Tiedemann and
	Reut Tsarfaty},
	title            = {First Align, then Predict: Understanding the Cross-Lingual Ability
	of Multilingual {BERT}},
	booktitle        = {Proceedings of the 16th Conference of the European Chapter of the
	Association for Computational Linguistics: Main Volume, {EACL} 2021,
	Online, April 19 - 23, 2021},
	pages            = {2214--2231},
	publisher        = {Association for Computational Linguistics},
	year             = {2021},
	burl             = {https://doi.org/10.18653/v1/2021.eacl-main.189},
	doi              = {10.18653/V1/2021.EACL-MAIN.189},
	bibsource        = {dblp computer science bibliography, https://dblp.org},
}

@inproceedings{libovicky2020language,
	author           = {Jindrich Libovick{\'{y}} and
	Rudolf Rosa and
	Alexander Fraser},
	editor           = {Trevor Cohn and
	Yulan He and
	Yang Liu},
	title            = {On the Language Neutrality of Pre-trained Multilingual Representations},
	booktitle        = {Findings of the Association for Computational Linguistics: {EMNLP}
	2020, Online Event, 16-20 November 2020},
	series           = {Findings of {ACL}},
	volume           = {{EMNLP} 2020},
	pages            = {1663--1674},
	publisher        = {Association for Computational Linguistics},
	year             = {2020},
	burl             = {https://doi.org/10.18653/v1/2020.findings-emnlp.150},
	doi              = {10.18653/V1/2020.FINDINGS-EMNLP.150},
	bibsource        = {dblp computer science bibliography, https://dblp.org},
}

@article{skirgard2023grambank,
	title            = {Grambank reveals the importance of genealogical constraints on linguistic diversity and highlights the impact of language loss},
	volume           = {9},
	issn             = {2375-2548},
	burl             = {http://dx.doi.org/10.1126/sciadv.adg6175},
	doi              = {10.1126/sciadv.adg6175},
	number           = {16},
	journal          = {Science Advances},
	publisher        = {American Association for the Advancement of Science (AAAS)},
	author           = {Skirgård, Hedvig and Haynie, Hannah J. and Blasi, Damián E. and Hammarström, Harald and Collins, Jeremy and Latarche, Jay J. and Lesage, Jakob and Weber, Tobias and Witzlack-Makarevich, Alena and Passmore, Sam and Chira, Angela and Maurits, Luke and Dinnage, Russell and Dunn, Michael and Reesink, Ger and Singer, Ruth and Bowern, Claire and Epps, Patience and Hill, Jane and Vesakoski, Outi and Robbeets, Martine and Abbas, Noor Karolin and Auer, Daniel and Bakker, Nancy A. and Barbos, Giulia and Borges, Robert D. and Danielsen, Swintha and Dorenbusch, Luise and Dorn, Ella and Elliott, John and Falcone, Giada and Fischer, Jana and Ghanggo Ate, Yustinus and Gibson, Hannah and Göbel, Hans-Philipp and Goodall, Jemima A. and Gruner, Victoria and Harvey, Andrew and Hayes, Rebekah and Heer, Leonard and Herrera Miranda, Roberto E. and Hübler, Nataliia and Huntington-Rainey, Biu and Ivani, Jessica K. and Johns, Marilen and Just, Erika and Kashima, Eri and Kipf, Carolina and Klingenberg, Janina V. and König, Nikita and Koti, Aikaterina and Kowalik, Richard G. A. and Krasnoukhova, Olga and Lindvall, Nora L. M. and Lorenzen, Mandy and Lutzenberger, Hannah and Martins, Tânia R. A. and Mata German, Celia and van der Meer, Suzanne and Montoya Samamé, Jaime and Müller, Michael and Muradoglu, Saliha and Neely, Kelsey and Nickel, Johanna and Norvik, Miina and Oluoch, Cheryl Akinyi and Peacock, Jesse and Pearey, India O. C. and Peck, Naomi and Petit, Stephanie and Pieper, Sören and Poblete, Mariana and Prestipino, Daniel and Raabe, Linda and Raja, Amna and Reimringer, Janis and Rey, Sydney C. and Rizaew, Julia and Ruppert, Eloisa and Salmon, Kim K. and Sammet, Jill and Schembri, Rhiannon and Schlabbach, Lars and Schmidt, Frederick W. P. and Skilton, Amalia and Smith, Wikaliler Daniel and de Sousa, Hilário and Sverredal, Kristin and Valle, Daniel and Vera, Javier and Voß, Judith and Witte, Tim and Wu, Henry and Yam, Stephanie and Ye, Jingting and Yong, Maisie and Yuditha, Tessa and Zariquiey, Roberto and Forkel, Robert and Evans, Nicholas and Levinson, Stephen C. and Haspelmath, Martin and Greenhill, Simon J. and Atkinson, Quentin D. and Gray, Russell D.},
	year             = {2023},
	month            = {apr},
}

@inproceedings{wals,
	author           = {Dryer, Matthew S. and Haspelmath, Martin},
	booktitle        = {WALS Online (v2020.4) [Data set]},
	year             = {2013},
	doi              = {10.5281/zenodo.13950591},
	url              = {https://wals.info},
	publisher        = {Zenodo},
	bnote             = {Data set. Accessed: 2026-02-20},
}

@article{evans2009myth,
	title            = {The myth of language universals: Language diversity and its importance for cognitive science},
	volume           = {32},
	issn             = {1469-1825},
	burl             = {http://dx.doi.org/10.1017/s0140525x0999094x},
	doi              = {10.1017/s0140525x0999094x},
	number           = {5},
	journal          = {Behavioral and Brain Sciences},
	publisher        = {Cambridge University Press (CUP)},
	author           = {Evans, Nicholas and Levinson, Stephen C.},
	year             = {2009},
	month            = {oct},
	pages            = {429–448},
}

@article{mitchell2008predicting,
	title            = {Predicting Human Brain Activity Associated with the Meanings of Nouns},
	volume           = {320},
	issn             = {1095-9203},
	burl             = {http://dx.doi.org/10.1126/science.1152876},
	doi              = {10.1126/science.1152876},
	number           = {5880},
	journal          = {Science},
	publisher        = {American Association for the Advancement of Science (AAAS)},
	author           = {Mitchell, Tom M. and Shinkareva, Svetlana V. and Carlson, Andrew and Chang, Kai-Min and Malave, Vicente L. and Mason, Robert A. and Just, Marcel Adam},
	year             = {2008},
	month            = {may},
	pages            = {1191–1195},
}

@article{nastase2021narratives,
	title   = {The ``{Narratives}'' {fMRI} dataset for evaluating models of
	naturalistic language comprehension},
	author  = {Nastase, Samuel A. and Liu, Yun-Fei and Hillman, Hanna
	and Zadbood, Asieh and Hasenfratz, Liat and Keshavarzian, Neggin
	and Chen, Janice and Honey, Christopher J. and Yeshurun, Yaara
	and Regev, Mor and Nguyen, Mai and Chang, Claire H. C.
	and Baldassano, Christopher and Lositsky, Olga and Simony, Erez
	and Chow, Michael A. and Leong, Yuan Chang and Brooks, Paula P.
	and Micciche, Emily and Choe, Gina and Goldstein, Ariel
	and Vanderwal, Tamara and Halchenko, Yaroslav O.
	and Norman, Kenneth A. and Hasson, Uri},
	journal = {Scientific Data},
	year    = {2021},
	volume  = {8},
	pages   = {250},
	doi     = {10.1038/s41597-021-01033-3},
	pmid    = {34584100},
}

@article{fedorenko2011functional,
	title            = {Functional specificity for high-level linguistic processing in the human brain},
	volume           = {108},
	issn             = {1091-6490},
	burl             = {http://dx.doi.org/10.1073/pnas.1112937108},
	doi              = {10.1073/pnas.1112937108},
	number           = {39},
	journal          = {Proceedings of the National Academy of Sciences},
	publisher        = {Proceedings of the National Academy of Sciences},
	author           = {Fedorenko, Evelina and Behr, Michael K. and Kanwisher, Nancy},
	year             = {2011},
	month            = {sep},
	pages            = {16428–16433},
}

@article{blank2017domain,
	author           = {Blank, Idan A. and Fedorenko, Evelina},
	title            = {Domain-General Brain Regions Do Not Track Linguistic Input as Closely as Language-Selective Regions},
	journal          = {The Journal of Neuroscience},
	year             = {2017},
	volume           = {37},
	number           = {41},
	pages            = {9999--10011},
	doi              = {10.1523/JNEUROSCI.3642-16.2017},
	url              = {https://doi.org/10.1523/JNEUROSCI.3642-16.2017},
	publisher        = {Society for Neuroscience},
}

@article{hickok2007cortical,
	title            = {The cortical organization of speech processing},
	volume           = {8},
	issn             = {1471-0048},
	burl             = {http://dx.doi.org/10.1038/nrn2113},
	doi              = {10.1038/nrn2113},
	number           = {5},
	journal          = {Nature Reviews Neuroscience},
	publisher        = {Springer Science and Business Media LLC},
	author           = {Hickok, Gregory and Poeppel, David},
	year             = {2007},
	month            = {apr},
	pages            = {393–402},
}

@article{perani2005neural,
	title            = {The neural basis of first and second language processing},
	volume           = {15},
	issn             = {0959-4388},
	burl             = {http://dx.doi.org/10.1016/j.conb.2005.03.007},
	doi              = {10.1016/j.conb.2005.03.007},
	number           = {2},
	journal          = {Current Opinion in Neurobiology},
	publisher        = {Elsevier BV},
	author           = {Perani, Daniela and Abutalebi, Jubin},
	year             = {2005},
	month            = {apr},
	pages            = {202–206},
}

@article{malikmoraleda2024polyglots,
	title   = {Functional characterization of the language network of polyglots
	and hyperpolyglots with precision {fMRI}},
	author  = {Malik-Moraleda, Saima and Jouravlev, Olessia and
	Taliaferro, Maya and Mineroff, Zachary and Cucu, Theodore
	and Mahowald, Kyle and Blank, Idan A. and Fedorenko, Evelina},
	journal = {Cerebral Cortex},
	year    = {2024},
	volume  = {34},
	number  = {3},
	doi     = {10.1093/cercor/bhae049},
	pmid    = {38466812},
}

@article{green2013language,
	title            = {Language control in bilinguals: The adaptive control hypothesis},
	volume           = {25},
	issn             = {2044-592X},
	burl             = {http://dx.doi.org/10.1080/20445911.2013.796377},
	doi              = {10.1080/20445911.2013.796377},
	number           = {5},
	journal          = {Journal of Cognitive Psychology},
	publisher        = {Informa UK Limited},
	author           = {Green, David W. and Abutalebi, Jubin},
	year             = {2013},
	month            = {may},
	pages            = {515–530},
}

@article{li2022petit,
	title            = {Le Petit Prince multilingual naturalistic fMRI corpus},
	volume           = {9},
	issn             = {2052-4463},
	burl             = {http://dx.doi.org/10.1038/s41597-022-01625-7},
	doi              = {10.1038/s41597-022-01625-7},
	number           = {1},
	journal          = {Scientific Data},
	publisher        = {Springer Science and Business Media LLC},
	author           = {Li, Jixing and Bhattasali, Shohini and Zhang, Shulin and Franzluebbers, Berta and Luh, Wen-Ming and Spreng, R. Nathan and Brennan, Jonathan R. and Yang, Yiming and Pallier, Christophe and Hale, John},
	year             = {2022},
	month            = {aug},
}

@article{momenian2024lpphk,
	title   = {Le {Petit Prince Hong Kong} ({LPPHK}): Naturalistic {fMRI} and
	{EEG} data from older {Cantonese} speakers},
	author  = {Momenian, Mohammad and Ma, Zhengwu and Wu, Shuyi
	and Wang, Chengcheng and Brennan, Jonathan and Hale, John
	and Meyer, Lars and Li, Jixing},
	journal = {Scientific Data},
	year    = {2024},
	volume  = {11},
	pages   = {992},
	doi     = {10.1038/s41597-024-03745-8},
	pmid    = {39261552},
}

@inproceedings{stehwien2020little,
	title = "The Little Prince in 26 Languages: Towards a Multilingual Neuro-Cognitive Corpus",
	author = "Stehwien, Sabrina  and
	Henke, Lena  and
	Hale, John  and
	Brennan, Jonathan  and
	Meyer, Lars",
	editor = "Chersoni, Emmanuele  and
	Devereux, Barry  and
	Huang, Chu-Ren",
	booktitle = "Proceedings of the Second Workshop on Linguistic and Neurocognitive Resources",
	month = may,
	year = "2020",
	address = "Marseille, France",
	publisher = "European Language Resources Association",
	url = "https://aclanthology.org/2020.lincr-1.6/",
	pages = "43--49",
	language = "eng",
	ISBN = "979-10-95546-52-8"
}

@article{siegelman2022meco,
	title            = {Expanding horizons of cross-linguistic research on reading: The Multilingual Eye-movement Corpus (MECO)},
	volume           = {54},
	issn             = {1554-3528},
	burl             = {http://dx.doi.org/10.3758/s13428-021-01772-6},
	doi              = {10.3758/s13428-021-01772-6},
	number           = {6},
	journal          = {Behavior Research Methods},
	publisher        = {Springer Science and Business Media LLC},
	author           = {Siegelman, Noam and Schroeder, Sascha and Acartürk, Cengiz and Ahn, Hee-Don and Alexeeva, Svetlana and Amenta, Simona and Bertram, Raymond and Bonandrini, Rolando and Brysbaert, Marc and Chernova, Daria and Da Fonseca, Sara Maria and Dirix, Nicolas and Duyck, Wouter and Fella, Argyro and Frost, Ram and Gattei, Carolina A. and Kalaitzi, Areti and Kwon, Nayoung and Lõo, Kaidi and Marelli, Marco and Papadopoulos, Timothy C. and Protopapas, Athanassios and Savo, Satu and Shalom, Diego E. and Slioussar, Natalia and Stein, Roni and Sui, Longjiao and Taboh, Analí and Tønnesen, Veronica and Usal, Kerem Alp and Kuperman, Victor},
	year             = {2022},
	month            = {feb},
	pages            = {2843–2863},
}

@inproceedings{millet2022toward,
	author           = {Juliette Millet and
	Charlotte Caucheteux and
	Pierre Orhan and
	Yves Boubenec and
	Alexandre Gramfort and
	Ewan Dunbar and
	Christophe Pallier and
	Jean{-}Remi King},
	editor           = {Sanmi Koyejo and
	S. Mohamed and
	A. Agarwal and
	Danielle Belgrave and
	K. Cho and
	A. Oh},
	title            = {Toward a realistic model of speech processing in the brain with self-supervised
	learning},
	booktitle        = {Advances in Neural Information Processing Systems 35: Annual Conference
	on Neural Information Processing Systems 2022, NeurIPS 2022, New Orleans,
	LA, USA, November 28 - December 9, 2022},
	year             = {2022},
	burl             = {http://papers.nips.cc/paper\_files/paper/2022/hash/d81ecfc8fb18e833a3fa0a35d92532b8-Abstract-Conference.html},
	bibsource        = {dblp computer science bibliography, https://dblp.org},
}

@book{snijders2012multilevel,
	author    = {Snijders, Tom A. B. and Bosker, Roel J.},
	title     = {Multilevel Analysis: An Introduction to Basic and
	Advanced Multilevel Modeling},
	edition   = {2nd},
	publisher = {SAGE},
	address   = {London},
	year      = {2012},
	isbn      = {978-1-84920-201-5},
}

@incollection{boas2012sign,
	author           = {Hans C. Boas and Ivan A. Sag},
	title            = {Sign-Based Construction Grammar},
	year             = {2012},
	publisher        = {CSLI Publications},
	address          = {Stanford, CA},
	series           = {CSLI Lecture Notes},
	number           = {193},
	isbn             = {978-1-57586-628-4},
}

@inbook{bergen2005embodied,
	title            = {Embodied Construction Grammar in simulation-based language understanding},
	isbn             = {9789027294708},
	issn             = {1573-594X},
	burl             = {http://dx.doi.org/10.1075/cal.3.08ber},
	doi              = {10.1075/cal.3.08ber},
	booktitle        = {Construction Grammars},
	publisher        = {John Benjamins Publishing Company},
	author           = {Bergen, Benjamin K. and Chang, Nancy},
	year             = {2005},
	month            = {feb},
	pages            = {147–190},
}

@article{berzak2022eye,
	title            = {CELER: A 365-Participant Corpus of Eye Movements in L1 and L2 English Reading},
	volume           = {6},
	issn             = {2470-2986},
	burl             = {http://dx.doi.org/10.1162/opmi_a_00054},
	doi              = {10.1162/opmi_a_00054},
	journal          = {Open Mind},
	publisher        = {MIT Press},
	author           = {Berzak, Yevgeni and Nakamura, Chie and Smith, Amelia and Weng, Emily and Katz, Boris and Flynn, Suzanne and Levy, Roger},
	year             = {2022},
	pages            = {41–50},
}

@article{qwen2025qwen25,
	author       = {An Yang and
	Baosong Yang and
	Beichen Zhang and
	Binyuan Hui and
	Bo Zheng and
	Bowen Yu and
	Chengyuan Li and
	Dayiheng Liu and
	Fei Huang and
	Haoran Wei and
	Huan Lin and
	Jian Yang and
	Jianhong Tu and
	Jianwei Zhang and
	Jianxin Yang and
	Jiaxi Yang and
	Jingren Zhou and
	Junyang Lin and
	Kai Dang and
	Keming Lu and
	Keqin Bao and
	Kexin Yang and
	Le Yu and
	Mei Li and
	Mingfeng Xue and
	Pei Zhang and
	Qin Zhu and
	Rui Men and
	Runji Lin and
	Tianhao Li and
	Tingyu Xia and
	Xingzhang Ren and
	Xuancheng Ren and
	Yang Fan and
	Yang Su and
	Yichang Zhang and
	Yu Wan and
	Yuqiong Liu and
	Zeyu Cui and
	Zhenru Zhang and
	Zihan Qiu},
	title        = {Qwen2.5 Technical Report},
	journal      = {arXiv preprint},
	volume       = {arXiv.2412.15115},
	year         = {2024},
	url          = {https://doi.org/10.48550/arXiv.2412.15115},
	doi          = {10.48550/ARXIV.2412.15115},
	beprinttype    = {arXiv},
	beprint       = {2412.15115},
	bibsource    = {dblp computer science bibliography, https://dblp.org}
}

@article{ostarek2024cross,
	author = {Martin Ulrich and Marcel Harpaintner and Natalie M. Trumpp and Alexander Berger and Fritz Günther and Markus Kiefer},
	doi = {10.1038/s41598-025-32189-2},
	issn = {20452322},
	issue = {1},
	journal = {Scientific Reports},
	title = {Indirect experiential grounding: semantic similarity of abstract scientific concepts is reflected in activity patterns in visual and motor cortex},
	volume = {15},
	year = {2025}
}

@inproceedings{rakshit2025constructions,
	title = "Meaning-infused grammar: Gradient Acceptability Shapes the Geometric Representations of Constructions in {LLM}s",
	author = "Rakshit, Supantho  and
	Goldberg, Adele E.",
	editor = "Bonial, Claire  and
	Torgbi, Melissa  and
	Weissweiler, Leonie  and
	Blodgett, Austin  and
	Beuls, Katrien  and
	Van Eecke, Paul  and
	Tayyar Madabushi, Harish",
	booktitle = "Proceedings of the Second International Workshop on Construction Grammars and NLP",
	month = sep,
	year = "2025",
	address = {D{\"u}sseldorf, Germany},
	publisher = "Association for Computational Linguistics",
	url = "https://aclanthology.org/2025.cxgsnlp-1.15/",
	pages = "151--157",
	ISBN = "979-8-89176-318-0"
}

@article{weissweiler2025llm,
	title={ReasonIF: Large Reasoning Models Fail to Follow Instructions During Reasoning}, 
	author={Yongchan Kwon and Shang Zhu and Federico Bianchi and Kaitlyn Zhou and James Zou},
	year={2025},
	journal      = {arXiv preprint},
	volume       = {arXiv.2510.15211},
	beprint={2510.15211},
	archivePrefix={arXiv},
	primaryClass={cs.LG},
	url={https://arxiv.org/arXiv.2510.15211}, 
}

@inproceedings{alkhamissi2025llm,
	
	title = "From Language to Cognition: How {LLM}s Outgrow the Human Language Network",
	author = "AlKhamissi, Badr  and
	Tuckute, Greta  and
	Tang, Yingtian  and
	Binhuraib, Taha Osama A  and
	Bosselut, Antoine  and
	Schrimpf, Martin",
	editor = "Christodoulopoulos, Christos  and
	Chakraborty, Tanmoy  and
	Rose, Carolyn  and
	Peng, Violet",
	booktitle = "Proceedings of the 2025 Conference on Empirical Methods in Natural Language Processing",
	month = nov,
	year = "2025",
	address = "Suzhou, China",
	publisher = "Association for Computational Linguistics",
	url = "https://aclanthology.org/2025.emnlp-main.1237/",
	doi = "10.18653/v1/2025.emnlp-main.1237",
	pages = "24321--24339",
	ISBN = "979-8-89176-332-6"
}

@article{gauthier2022does,
	author = {Ariel Goldstein and Eric Ham and Mariano Schain and Samuel A. Nastase and Bobbi Aubrey and Zaid Zada and Avigail Grinstein-Dabush and Harshvardhan Gazula and Amir Feder and Werner Doyle and Sasha Devore and Patricia Dugan and Daniel Friedman and Michael Brenner and Avinatan Hassidim and Yossi Matias and Orrin Devinsky and Noam Siegelman and Adeen Flinker and Omer Levy and Roi Reichart and Uri Hasson},
	doi = {10.1038/s41467-025-65518-0},
	issn = {20411723},
	issue = {1},
	journal = {Nature Communications},
	title = {Temporal structure of natural language processing in the human brain corresponds to layered hierarchy of large language models},
	volume = {16},
	year = {2025}
}

\appendix

% ============================================
% APPENDIX A: Conceptual Framework
% ============================================
\section{Conceptual Framework}
\label{sec:appendix_framework}

\begin{figure*}[t]
	\centering
	\resizebox{\textwidth}{!}{
		\begin{tikzpicture}[
			box/.style={draw, rounded corners=4pt, minimum height=0.9cm, align=center, font=\small, line width=0.4pt},
			bigbox/.style={draw, rounded corners=6pt, minimum height=1.15cm, align=center, font=\small, line width=0.6pt},
			pipebox/.style={draw, rounded corners=4pt, minimum height=0.9cm, align=center, font=\small, line width=0.4pt},
			arrow/.style={-{Stealth[length=2.5mm, width=1.8mm]}, line width=0.5pt},
			thinnarrow/.style={-{Stealth[length=2mm, width=1.5mm]}, line width=0.4pt},
			lbl/.style={font=\scriptsize\itshape, text=gray!55!black}
			]
			
			% ===== ROW 1: Top-level boxes =====
			\node[bigbox, fill=blue!8, draw=blue!40, text width=2.8cm] at (0, 0) (brain) 
			{Universal Language\\Network\\{\scriptsize (fMRI, $N\!=\!112$)}};
			
			\node[bigbox, fill=green!8, draw=green!50!black, text width=2.8cm] at (6.8, 0) (llm) 
			{LLM\\Representations\\{\scriptsize (7 models)}};
			
			\node[pipebox, fill=yellow!8, draw=yellow!50!black, text width=1.7cm] at (11.0, 0) (pca) 
			{PCA\\{\scriptsize ($d\!\to\!100$)}};
			
			\node[pipebox, fill=yellow!8, draw=yellow!50!black, text width=1.7cm] at (13.7, 0) (ridge) 
			{Ridge\\Regression};
			
			\node[pipebox, fill=purple!8, draw=purple!40, text width=1.7cm] at (16.4, 0) (eval) 
			{$r$, $\tilde{r}$\\{\scriptsize per ROI}};
			
			% Pipeline arrows
			\draw[arrow, black!60] (llm) -- (pca);
			\draw[arrow, black!60] (pca) -- (ridge);
			\draw[arrow, black!60] (ridge) -- (eval);
			
			% Dashed brain -> ridge (routes above pipeline)
			\draw[arrow, blue!50, dashed, line width=0.5pt]
			(brain.north east) -- ++(0.6, 0.5)
			node[pos=0.6, above, lbl] {$y_v$}
			-- ([yshift=8mm]ridge.north) -- (ridge.north);
			
			% ===== ROW 2: Language boxes =====
			\node[box, fill=red!8, draw=red!30, text width=1.5cm] at (-2.3, -2.2) (en) 
			{English\\{\scriptsize $n\!=\!49$}};
			\node[box, fill=red!8, draw=red!30, text width=1.5cm] at (0, -2.2) (zh) 
			{Chinese\\{\scriptsize $n\!=\!35$}};
			\node[box, fill=red!8, draw=red!30, text width=1.5cm] at (2.3, -2.2) (fr) 
			{French\\{\scriptsize $n\!=\!28$}};
			
			\draw[thinnarrow, blue!50] (brain) -- (en);
			\draw[thinnarrow, blue!50] (brain) -- (zh);
			\draw[thinnarrow, blue!50] (brain) -- (fr);
			
			% ===== ROW 2: Model boxes =====
			\node[box, fill=green!12, draw=green!40!black, text width=2.0cm, font=\scriptsize] at (4.6, -2.2) (en_dom) 
			{EN-dominant\\GPT-2, LLaMA-2};
			\node[box, fill=orange!12, draw=orange!50!black, text width=1.5cm, font=\scriptsize] at (6.8, -2.2) (zh_dom) 
			{ZH-dominant\\Baichuan2};
			\node[box, fill=green!20, draw=green!40!black, text width=2.2cm, font=\scriptsize] at (9.1, -2.2) (multi) 
			{Multilingual\\mBERT, XLM-R,\\BLOOM, Qwen2.5};
			
			\draw[thinnarrow, green!50!black] (llm) -- (en_dom);
			\draw[thinnarrow, orange!60!black] (llm) -- (zh_dom);
			\draw[thinnarrow, green!50!black] (llm) -- (multi);
			
			% ===== ROW 3: Predictions =====
			\node[font=\small\bfseries] at (7, -4.0) (rq) {
				\textcolor{orange!70!black}{P1: Universal alignment?} \qquad
				\textcolor{blue!60!black}{P2: Syntax--semantics ROI gradient?} \qquad
				\textcolor{green!50!black}{P3: Training dominance?} \qquad
				\textcolor{red!60!black}{P4: Surprisal mediation?}
			};
			
	\end{tikzpicture}}
	\caption{Conceptual framework. Four theory-derived predictions tested across three languages and seven models with noise-ceiling normalization.}
	\label{fig:framework}
\end{figure*}
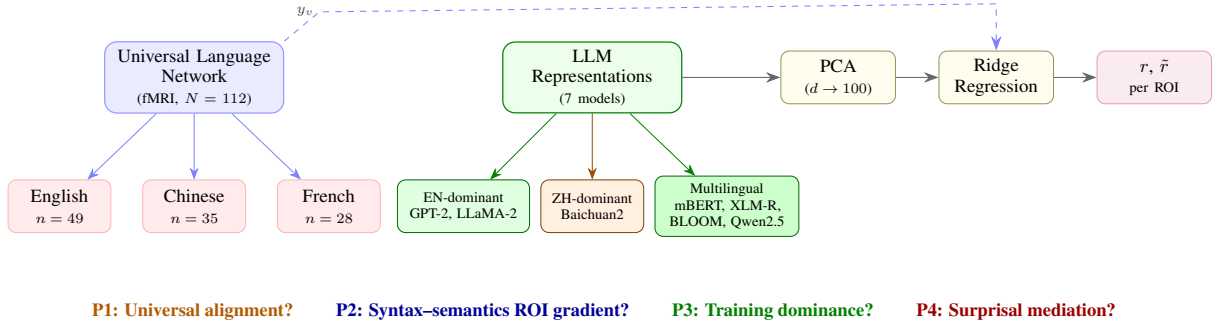

Figure~\ref{fig:framework} summarizes our experimental setup: fMRI data from three languages and seven LLMs are fed through identical PCA + ridge-regression encoding pipelines, with each of four pre-registered predictions (P1--P4) mapped onto a specific contrast or covariate in the resulting alignment matrix.

% ============================================
% APPENDIX B: Full SE Table
% ============================================
\section{Encoding Performance Standard Errors}
\label{sec:appendix_se}

Table~\ref{tab:appendix_se} reports the standard error of the mean encoding correlation for each model--language pair, computed as the across-subject standard deviation of voxel-mean correlations divided by $\sqrt{n_\text{subj}}$. All SEs lie below 0.015, indicating that the cross-language gradients summarized in Table~\ref{tab:main_results} are not driven by per-subject noise.

\begin{table}[h]
    \centering
    \small
    \begin{tabular}{lccc}
        \toprule
        \textbf{Model} & \textbf{SE(EN)} & \textbf{SE(ZH)} & \textbf{SE(FR)} \\
        \midrule
        GPT-2       & .011 & .009 & .012 \\
        LLaMA-2-7B  & .013 & .010 & .014 \\
        Baichuan2-7B & .012 & .012 & .011 \\
        mBERT       & .010 & .011 & .013 \\
        XLM-R       & .014 & .013 & .012 \\
        BLOOM-7B    & .013 & .012 & .013 \\
        Qwen2.5-7B  & .012 & .011 & .014 \\
        \bottomrule
    \end{tabular}
    \caption{Standard errors of mean encoding correlations ($r$) for each model--language combination. All SEs $< .015$.}
    \label{tab:appendix_se}
\end{table}

% ============================================
% APPENDIX C: Full ROI Results
% ============================================
\section{Full ROI Results}
\label{sec:appendix_roi}

Table~\ref{tab:appendix_roi} reports noise-ceiling-normalized encoding ($\tilde{r}$) by ROI for three representative models (XLM-R, Baichuan2-7B, LLaMA-2-7B). Two patterns are visible. First, PTL shows the highest alignment in every model--language cell, consistent with the lexico-semantic gradient \citep{kauf2024lexical}. Second, the across-language drop for language-dominant models is concentrated in the more anterior ROIs (IFG, MFG), foreshadowing the region-specific gradient analysis of \S\ref{sec:results} (Experiment~5).

\begin{table*}[t]
    \centering
    \small
    \begin{tabular}{llcccccc}
        \toprule
        \textbf{Model} & \textbf{Lang} & \textbf{IFG} & \textbf{MFG} & \textbf{ATL} & \textbf{PTL} & \textbf{AG} & \textbf{TP} \\
        \midrule
        XLM-R & EN & .78 & .81 & .85 & .89 & .82 & .79 \\
        XLM-R & ZH & .64 & .74 & .81 & .88 & .79 & .76 \\
        XLM-R & FR & .73 & .78 & .83 & .87 & .80 & .77 \\
        \midrule
        Baichuan2 & EN & .41 & .52 & .58 & .68 & .55 & .49 \\
        Baichuan2 & ZH & .79 & .83 & .87 & .90 & .84 & .81 \\
        Baichuan2 & FR & .37 & .48 & .54 & .63 & .51 & .46 \\
        \midrule
        LLaMA-2 & EN & .80 & .83 & .87 & .91 & .84 & .81 \\
        LLaMA-2 & ZH & .32 & .43 & .49 & .59 & .46 & .41 \\
        LLaMA-2 & FR & .58 & .65 & .71 & .78 & .68 & .63 \\
        \bottomrule
    \end{tabular}
    \caption{Noise-ceiling-normalized encoding ($\tilde{r}$) by ROI for selected models. PTL shows the highest alignment across all conditions. Note: Baichuan2 Chinese ATL ($\tilde{r} = .87$) and PTL ($\tilde{r} = .90$) approach near-ceiling performance; the near-ceiling values should be interpreted cautiously as they may be influenced by noise ceiling estimation uncertainty.}
    \label{tab:appendix_roi}
\end{table*}

% ============================================
% APPENDIX D: Layer-Wise Details
% ============================================
\section{Layer-Wise Details}
\label{sec:appendix_layers}

The Chinese rightward shift replicates across multiple models: BLOOM-7B (peak shift: +1.5 layers, $d = 0.85$); mBERT (peak shift: +1.0 layers, $d = 0.61$); LLaMA-2-7B (peak shift: +2.1 layers, $d = 1.04$). The Baichuan2 peak layer for Chinese is \emph{earlier} than for English (18.1 vs.\ 19.8), the reverse of the pattern seen with other models, consistent with Baichuan2's low Chinese fertility (1.6 vs.\ 3.8 for LLaMA-2) and further supporting the tokenization-mediated account.

% ============================================
% APPENDIX E: PCA Variance Explained
% ============================================
\section{Tokenization Fertility}
\label{sec:appendix_fertility}

Table~\ref{tab:fertility} reports tokenization fertility (the average number of subword tokens produced per orthographic word in the LPP stimulus text) across every language $\times$ tokenizer combination used in our experiments. Two regularities are visible. Tokenizers trained predominantly on English produce nearly word-level segmentation for English and French but break Chinese into many more pieces (1.7--2.9$\times$). Tokenizers with substantial Chinese training (Baichuan2, Qwen2.5) achieve near-parity across all three languages, demonstrating that the Chinese-high-fertility pattern is a property of vocabulary composition rather than of Chinese script per se.

\begin{table}[t]
    \centering
    \small
    \begin{tabular}{lccc}
        \toprule
        \textbf{Tokenizer} & \textbf{EN} & \textbf{ZH} & \textbf{FR} \\
        \midrule
        LLaMA-2      & 1.3 & 3.8 & 1.5 \\
        Baichuan2    & 1.4 & 1.6 & 1.6 \\
        XLM-R SentencePiece & 1.4 & 2.4 & 1.5 \\
        mBERT WordPiece & 1.2 & 2.1 & 1.4 \\
        BLOOM BPE    & 1.3 & 2.6 & 1.4 \\
        Qwen2.5 BPE  & 1.3 & 1.8 & 1.5 \\
        \bottomrule
    \end{tabular}
    \caption{Tokenization fertility: mean subword tokens per word in the LPP stimulus text. Chinese shows consistently higher fertility except for Baichuan2 and Qwen2.5, whose vocabularies are optimized for Chinese.}
    \label{tab:fertility}
\end{table}

\section{PCA Variance Explained}
\label{sec:appendix_pca}

Table~\ref{tab:appendix_pca_var} reports cumulative variance explained by the top 50, 100, and 200 principal components of each model's representations (best-performing layer, English). The 100-PC choice used throughout the main analysis retains 85--93\% of variance across all models. A robustness analysis varying the cut at 50, 100, 150, and 200 PCs produced encoding-correlation differences $\Delta r < 0.008$, ruling out PCA bandwidth as a meaningful source of cross-language variation.

\begin{table}[t]
    \centering
    \small
    \begin{tabular}{lccc}
        \toprule
        \textbf{Model} & \textbf{50 PCs} & \textbf{100 PCs} & \textbf{200 PCs} \\
        \midrule
        GPT-2       & 78\% & 89\% & 96\% \\
        LLaMA-2-7B  & 72\% & 85\% & 94\% \\
        Baichuan2-7B & 74\% & 87\% & 95\% \\
        mBERT       & 81\% & 93\% & 98\% \\
        XLM-R       & 76\% & 88\% & 95\% \\
        BLOOM-7B    & 75\% & 86\% & 94\% \\
        Qwen2.5-7B  & 73\% & 86\% & 95\% \\
        \bottomrule
    \end{tabular}
    \caption{Variance explained by PCA across models (best layer, English). 100 PCs explain 85--93\% of variance. Robustness analysis (50/100/150/200 PCs) yields stable encoding results ($\Delta r < .008$).}
    \label{tab:appendix_pca_var}
\end{table}

% ============================================
% APPENDIX F: Individual Variability
% ============================================
\section{Individual Variability}
\label{sec:appendix_individual}

Table~\ref{tab:appendix_individual} reports the across-subject distribution of XLM-R encoding correlations within each language. Most subjects (47/49 EN, 33/35 ZH, 27/28 FR) show above-chance encoding, indicating that aggregate cross-language patterns are not produced by a small subset of outlier subjects. The medians (.19--.21) are tightly clustered across languages, while the broader spread for English likely reflects its larger sample size rather than language-specific variability.

\begin{table}[t]
    \centering
    \small
    \begin{tabular}{lccccc}
        \toprule
        \textbf{Lang} & \textbf{Min} & \textbf{Q1} & \textbf{Med} & \textbf{Q3} & \textbf{Max} \\
        \midrule
        EN & .09 & .18 & .21 & .24 & .31 \\
        ZH & .05 & .15 & .19 & .23 & .29 \\
        FR  & .08 & .16 & .20 & .23 & .28 \\
        \bottomrule
    \end{tabular}
    \caption{Distribution of individual-subject encoding correlations (XLM-R). 47/49 EN, 33/35 ZH, 27/28 FR subjects show above-chance encoding. Individual-difference analyses (e.g., L2 proficiency, language exposure) could reveal whether the typological gradient varies with individual linguistic experience; we reserve this for future work.}
    \label{tab:appendix_individual}
\end{table}

% ============================================
% APPENDIX H: Typological Distance Details
% ============================================
\section{Typological Distance Details}
\label{sec:appendix_typology}

Table~\ref{tab:typological} reports pairwise typological distances under three metrics alongside the alignment drop (in percent of raw $r$) each model exhibits when tested on a language other than its training-dominant language. The three metrics agree closely on the rank-ordering EN--FR $<$ FR--ZH $<$ EN--ZH, supporting the use of any of them as the distance covariate; Grambank is preferred in the main analysis because of its dense, balanced feature set. The alignment-drop rows show the central asymmetry: both language-dominant models degrade steeply with typological distance, while the multilingual model (XLM-R) shows only minor drops.

\begin{table}[t]
	\centering
	\small
	\setlength{\tabcolsep}{4pt}
	\begin{tabular}{@{}l ccc@{}}
		\toprule
		& \textbf{EN--FR} & \textbf{EN--ZH} & \textbf{FR--ZH} \\
		\midrule
		Grambank    & 0.21 & 0.48 & 0.44 \\
		WALS        & 0.18 & 0.52 & 0.47 \\
		lang2vec (syn.) & 0.24 & 0.61 & 0.55 \\
		\midrule
		\multicolumn{4}{@{}l}{\emph{Alignment drop from dominant lang.\ (\%)}} \\
		LLaMA-2 (EN-dom.)    & $-$20 & $-$45 & --- \\
		Baichuan2 (ZH-dom.)  & --- & --- & $-$34 \\
		XLM-R (multilin.)    & $-$5  & $-$8  & --- \\
		\bottomrule
	\end{tabular}
	\caption{Typological distances and corresponding alignment drops (\%) from each model's dominant training language. Drops are computed from raw $r$ values.}
	\label{tab:typological}
\end{table}

%% ---- FIGURE: Typological distance with 3 model types ---- %%
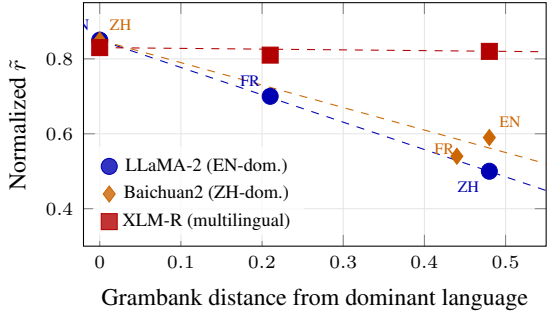
\begin{figure}[t]
\centering
\begin{tikzpicture}
\begin{axis}[
    width=\columnwidth,
    height=4.8cm,
    xlabel={Grambank distance from dominant language},
    ylabel={Normalized $\tilde{r}$},
    xmin=-0.02, xmax=0.55,
    ymin=0.3, ymax=0.95,
    legend style={at={(0.02,0.02)}, anchor=south west, font=\scriptsize, draw=none, fill=white, fill opacity=0.8, text opacity=1},
    grid=major,
    grid style={gray!20},
    tick label style={font=\scriptsize},
    label style={font=\small},
]

% LLaMA-2 (English-dominant)
\addplot[only marks, mark=*, mark size=3pt, blue!70!black] coordinates {
    (0, 0.85) (0.21, 0.70) (0.48, 0.50)
};
\addplot[blue!70!black, dashed, domain=0:0.55, samples=2] {0.85 - 0.73*x};

% Baichuan2 (Chinese-dominant) -- distances from ZH
\addplot[only marks, mark=diamond*, mark size=3pt, orange!80!black] coordinates {
    (0, 0.85) (0.44, 0.54) (0.48, 0.59)
};
\addplot[orange!80!black, dashed, domain=0:0.55, samples=2] {0.85 - 0.60*x};

% XLM-R (Multilingual)
\addplot[only marks, mark=square*, mark size=3pt, red!70!black] coordinates {
    (0, 0.83) (0.21, 0.81) (0.48, 0.82)
};
\addplot[red!70!black, dashed, domain=0:0.55, samples=2] {0.83 - 0.02*x};

\node[font=\tiny, blue!70!black, anchor=south east] at (axis cs:0,0.85) {EN};
\node[font=\tiny, blue!70!black, anchor=south east] at (axis cs:0.21,0.70) {FR};
\node[font=\tiny, blue!70!black, anchor=north east] at (axis cs:0.48,0.50) {ZH};

\node[font=\tiny, orange!80!black, anchor=south west] at (axis cs:0,0.85) {ZH};
\node[font=\tiny, orange!80!black, anchor=north west] at (axis cs:0.4,0.6) {FR};
\node[font=\tiny, orange!80!black, anchor=south west] at (axis cs:0.48,0.59) {EN};

\legend{LLaMA-2 (EN-dom.), , Baichuan2 (ZH-dom.), , XLM-R (multilingual)}
\end{axis}
\end{tikzpicture}
\caption{Encoding performance ($\tilde{r}$) vs.\ Grambank distance from each model's dominant training language. Dashed lines: linear fits. Both language-dominant models show steep degradation; XLM-R (multilingual) is flat. Baichuan2 distances are computed from Chinese (its dominant language); LLaMA-2 distances from English.}
\label{fig:typological}
\end{figure}

\begin{table}[t]
    \centering
    \small
    \begin{tabular}{lccc}
        \toprule
        \textbf{Metric} & \textbf{EN--FR} & \textbf{EN--ZH} & \textbf{FR--ZH} \\
        \midrule
        Grambank (Hamming) & 0.21 & 0.48 & 0.44 \\
        WALS (Hamming)     & 0.18 & 0.52 & 0.47 \\
        lang2vec syntactic & 0.24 & 0.61 & 0.55 \\
        lang2vec phonological & 0.31 & 0.58 & 0.53 \\
        Genealogical       & 0.33 & 1.00 & 1.00 \\
        Geographic (norm.) & 0.08 & 0.72 & 0.68 \\
        \bottomrule
    \end{tabular}
    \caption{Full pairwise typological, genealogical, and geographic distances (normalized to [0, 1]).}
    \label{tab:appendix_distances}
\end{table}

\begin{table}[t]
    \centering
    \small
    \begin{tabular}{lccccc}
        \toprule
        \textbf{Model} & \textbf{EN}~$\tilde{r}$ & \textbf{ZH}~$\tilde{r}$ & \textbf{FR}~$\tilde{r}$ & $U$ & $U$~CI \\
        \midrule
        LLaMA-2-7B  & .85 & .50 & .70 & .71 & (.63,.79) \\
        Baichuan2-7B & .59 & .85 & .54 & .78 & (.70,.85) \\
        BLOOM-7B    & .78 & .78 & .77 & .96 & (.93,.98) \\
        \bottomrule
    \end{tabular}
    \caption{Autoregressive-only comparison (all 7B). BLOOM achieves dramatically higher uniformity than either language-dominant model. $U$ = uniformity (Eq.~\ref{eq:uniformity}); 95\% bootstrap CIs (10,000 resamples). $U$ is computed from raw $r$ values using sample standard deviation.}
    \label{tab:autoregressive}
\end{table}

% ============================================
% APPENDIX I: Sensitivity Analysis
% ============================================
\section{Sensitivity Analysis for Baichuan2 Training Proportion}
\label{sec:appendix_sensitivity}

Because Baichuan2's training-language composition is reported as approximately 55\% Chinese without an exact split per source-language, we re-estimated the partial-correlation and slope coefficients under values of $p_{\text{ZH}}$ ranging from 40\% to 70\% in 5-point increments (Table~\ref{tab:appendix_sensitivity}). The typological-distance coefficient $\beta$ and partial correlation vary by less than 0.07 across this 30-percentage-point range, and all qualitative conclusions are preserved. This bounds the influence of imprecision in our training-proportion covariate.

\begin{table}[t]
    \centering
    \small
    \begin{tabular}{lccc}
        \toprule
        \textbf{Assumed ZH \%} & $r_{\text{partial}}$ & $\beta$ & \textbf{Conclusion} \\
        \midrule
        40\% & $-0.31$ & $-0.38$ & Stable \\
        45\% & $-0.32$ & $-0.39$ & Stable \\
        50\% & $-0.33$ & $-0.40$ & Stable \\
        55\% (baseline) & $-0.34$ & $-0.41$ & Stable \\
        60\% & $-0.35$ & $-0.42$ & Stable \\
        65\% & $-0.36$ & $-0.43$ & Stable \\
        70\% & $-0.37$ & $-0.44$ & Stable \\
        \bottomrule
    \end{tabular}
    \caption{Sensitivity analysis: partial correlation (alignment $\sim$ typological distance $|$ training proportion) and $\beta$ under varying assumptions about Baichuan2's Chinese training proportion. All results remain qualitatively stable across the full $\pm 15$\% range. The near-linear increments reflect the approximately linear relationship between the training-proportion covariate and the partial correlation over this narrow range; exact values may show minor deviations at extremes depending on convergence criteria.}
    \label{tab:appendix_sensitivity}
\end{table}

% ============================================
% APPENDIX J: Cross-Language Transfer (Additional Models)
% ============================================
\section{Cross-Language Transfer: Additional Models}
\label{sec:appendix_transfer}

Table~\ref{tab:appendix_transfer_extra} summarizes cross-language encoding transfer for the two language-dominant 7B models. LLaMA-2 retains $\sim$61\% of within-language performance on its typologically closer non-dominant language (French) and $\sim$36\% on its more distant non-dominant language (Chinese). Baichuan2 shows the mirror-image pattern (the typologically closer non-dominant language being French at distance 0.44, the more distant being English at 0.48, with comparable gradient magnitudes), further supporting the training-dominance account introduced in \S\ref{sec:confound}.

\begin{table}[t]
    \centering
    \small
    \begin{tabular}{lcc}
        \toprule
        \textbf{Train $\to$ Test} & \textbf{LLaMA-2} $\tilde{r}$ & \textbf{Baichuan2} $\tilde{r}$ \\
        \midrule
        Dominant $\to$ Same  & .85 & .85 \\
        Dominant $\to$ Close & .52 & .41 \\
        Dominant $\to$ Far   & .31 & .36 \\
        \bottomrule
    \end{tabular}
    \caption{Cross-language transfer for language-dominant models. ``Close'' = typologically closer non-dominant language; ``Far'' = typologically distant non-dominant language. Baichuan2 shows a reversed gradient paralleling the within-language alignment pattern.}
    \label{tab:appendix_transfer_extra}
\end{table}

% ============================================
% APPENDIX K: Bonferroni Correction Results
% ============================================
\section{Multiple Comparison Correction}
\label{sec:appendix_holm}

Table~\ref{tab:appendix_holm} reports Bonferroni-corrected $p$-values for our four pre-registered predictions (\S\ref{sec:rep_vs_comp}). All four remain significant after correction. We use Bonferroni rather than Holm here because the four predictions were specified before data inspection and we treat them as independent confirmatory tests.

\begin{table}[t]
	\centering
	\small
	\begin{tabular}{@{}lccl@{}}
		\toprule
		\textbf{Prediction} & $p_{\text{uncorr}}$ & $p_{\text{Bonf}}$ & \textbf{Result} \\
		\midrule
		P1 (Universality)  & $< .001$ & $< .004$ & Confirmed \\
		P2 (ROI gradient)  & $< .01$  & $.03$    & Supported \\
		P3 (Training dom.) & $< .001$ & $< .001$ & Confirmed \\
		P4 (Surprisal)     & $< .01$  & $.03$    & Part.\ supported \\
		\bottomrule
	\end{tabular}
	\caption{Bonferroni correction across the four pre-registered predictions. All predictions remain significant after correction.}
	\label{tab:appendix_holm}
\end{table}

% ============================================
% APPENDIX L: Whole-Brain Analysis
% ============================================
\section{Whole-Brain Analysis}
\label{sec:appendix_wholebrain}

Whole-brain encoding reveals significant effects in the default mode network (DMN), including medial prefrontal cortex and posterior cingulate, consistent with narrative comprehension. The cross-linguistic pattern in DMN parallels language network results: multilingual models show more uniform performance. Absolute magnitude is lower (mean $r = 0.11$ vs.\ $0.19$ in language network for XLM-R). The MD network shows minimal encoding ($r < 0.04$, n.s.).

% ============================================
% APPENDIX M: Cross-Language Transfer (Full Results)
% ============================================
\section{Cross-Language Transfer: Full Results}
\label{sec:appendix_transfer_main}

Table~\ref{tab:transfer} reports the full set of cross-language transfer results for XLM-R: encoding models trained on subjects listening to one language and tested on subjects listening to another. Discussion of the patterns follows the table.

\begin{table}[t]
    \centering
    \small
    \begin{tabular}{lcccc}
        \toprule
        \textbf{Train $\to$ Test} & $r$ & $\tilde{r}$ & \textbf{\% within} & $p$ \\
        \midrule
        \multicolumn{5}{l}{\emph{XLM-R}} \\
        EN $\to$ EN  & .208 & .83 & 100\% & --- \\
        EN $\to$ FR  & .151 & .62 & 73\%  & $<.001$ \\
        EN $\to$ ZH  & .109 & .47 & 52\%  & $<.001$ \\
        ZH $\to$ ZH  & .191 & .82 & 100\% & --- \\
        ZH $\to$ EN  & .118 & .47 & 57\%  & $<.001$ \\
        ZH $\to$ FR  & .102 & .42 & 53\%  & $<.001$ \\
        FR $\to$ FR  & .197 & .81 & 100\% & --- \\
        FR $\to$ EN  & .138 & .55 & 66\%  & $<.001$ \\
        FR $\to$ ZH  & .098 & .42 & 50\%  & $<.001$ \\
        \bottomrule
    \end{tabular}
    \caption{Cross-language encoding transfer (XLM-R) with noise-ceiling-normalized $\tilde{r}$. Transfer to typologically closer pairs (EN$\to$FR: 73\%) outperforms distant pairs (FR$\to$ZH: 50\%). LLaMA-2 and Baichuan2 transfer results are reported in Appendix~\ref{sec:appendix_transfer}. The \emph{\% within} column normalizes raw transfer $r$ against the within-language baseline of the language whose tokenizer dominates the encoding (English's $r = 0.208$ for transfers involving English; the source language's within-language $r$ otherwise), which captures the practical question of how much of the achievable encoding ceiling is recoverable across the language boundary.}
    \label{tab:transfer}
\end{table}

All cross-language transfer results are significantly above chance ($p < 0.001$). Several patterns emerge. First, the transfer gradient aligns with typological distance: EN$\to$FR transfer retains 73\% of within-language performance, substantially exceeding EN$\to$ZH (52\%) and FR$\to$ZH (50\%). Second, the asymmetry between ZH$\to$EN (57\%) and ZH$\to$FR (53\%) may reflect the higher proportion of English in XLM-R's training data rather than typological proximity, since Grambank distance places Chinese slightly closer to French (0.44) than to English (0.48). Third, FR$\to$EN (66\%) exceeds both FR$\to$ZH (50\%) and EN$\to$FR (62\%), consistent with the Indo-European pair sharing more transferable representations. These patterns reinforce the finding that typological distance, training data composition, and genealogical relatedness all contribute to cross-linguistic brain--LLM alignment.

% ============================================
% APPENDIX N: CxG Specificity — Extended Discussion
% ============================================
\section{Future Predictions for $\geq$10 Languages}
\label{sec:appendix_future_predictions}

Our results generate three specific predictions for future cross-linguistic work: (a) typological features indexing structural variation (e.g., head direction, topic-prominence, morphological complexity) should predict ROI-specific alignment gradients more strongly than features indexing semantic organization; (b) the IFG gradient should be steeper for language pairs differing in argument-structure properties than for pairs differing primarily in phonological features; and (c) agglutinative languages (e.g., Turkish, Finnish) should show a layer shift in the opposite direction from Chinese if the shift reflects morphological decomposition demands rather than information density.

% ============================================
% APPENDIX O: CxG Specificity — Extended Discussion
% ============================================
\section{CxG Specificity: Extended Analysis}
\label{sec:appendix_cxg}

\subsection{Detailed Alternative Accounts}

Three alternative accounts predict the same IFG $>$ PTL typological gradient pattern observed in our data:

\paragraph{Generic syntax--semantics dissociation.} Any theory that posits cross-linguistic syntactic variation while maintaining relative semantic universality (including generative grammar and functionalist approaches) would predict larger typological effects in syntax-associated regions \citep{mahowald2024dissociating}. Under this account, IFG's steeper gradient simply reflects the well-established cross-linguistic variability of syntactic structure relative to semantic content, without requiring construction-specific representations.

\paragraph{Cognitive control.} IFG is implicated in cognitive control for non-dominant language processing \citep{green2013language}. The steeper IFG gradient could partially reflect increased control demands when LLM representations diverge from the listener's native language, rather than constructional processing per se. If this account is correct, the gradient should correlate with measures of cognitive control demand (e.g., conflict adaptation, switching costs) rather than purely linguistic typological distance.

\paragraph{Signal-to-noise ratio.} \citet{kauf2024lexical} showed that lexical-semantic content dominates brain--LLM similarity. Because IFG contributes less to overall alignment than PTL, its signal may be noisier and more susceptible to degradation with typological distance, producing a steeper apparent gradient without implicating constructional specificity. Under this account, normalizing for baseline alignment magnitude should attenuate the IFG/PTL gradient difference.

\subsection{What CxG-Specific Evidence Would Require}

A genuinely CxG-specific test would need to go beyond ROI-level gradients to examine construction-level alignment. Three approaches would provide uniquely CxG evidence:

\paragraph{Construction-type-specific analysis.} Testing whether sentences containing Chinese b\v{a}-constructions versus English double-object constructions produce different IFG alignment patterns. CxG predicts that specific form--function mappings, not just generic syntactic complexity, drive the gradient. This requires construction-annotated stimuli across languages.

\paragraph{Gradient category boundaries.} CxG predicts gradient, not categorical, constructional differences across languages, the neural gradient should itself be gradient rather than step-like. Languages with partially overlapping constructional inventories (e.g., English and French datives) should show intermediate gradients compared to languages with fully different constructions (e.g., English and Chinese disposal constructions).

\paragraph{Form--function dissociation.} Constructions that are form-similar but function-different across languages should show different alignment patterns than form-different but function-similar pairs. For example, English and French passive constructions share similar form but may differ in discourse function; if the IFG gradient tracks functional rather than formal similarity, this would support CxG over purely formal syntactic accounts.

We note that the Universal Constructicon (UCxn) project's 10-language annotations \citep{boas2012sign} could provide the infrastructure for such tests. The present finding is best described as consistent with, but not uniquely supporting, CxG.

% ============================================
% APPENDIX O: Ethical Considerations
% ============================================
% ============================================
% APPENDIX O: Ethical Considerations
% ============================================
\section{Ethical Considerations}
\label{sec:appendix_ethics}

The LPP corpus \citep{li2022petit} is publicly available on OpenNeuro (ds003643) under the CC0 license. The original study obtained IRB approval from Cornell University, Jiangsu Normal University, and NeuroSpin; all participants provided informed written consent. Participants were recruited from university communities and compensated at standard institutional rates. Our study involves secondary analysis of de-identified data and does not require additional IRB approval. All models are publicly available under open-source or research-use licenses. Code for all analyses is publicly released at \url{https://github.com/bettyguo/cross-lingual-brain-llm}.

\end{document}